\documentclass[1p, times]{elsarticle}
\usepackage{subeqnarray}
\usepackage{amsmath}
\usepackage{amssymb}
\usepackage[boxed]{algorithm2e}
\usepackage{multirow,booktabs}
\usepackage{graphicx,subfigure}
\usepackage{url}
\usepackage[normalem]{ulem}
\usepackage{fancybox}

\title{Feature Selection for MAUC-Oriented Classification Systems}
\author{Rui Wang}
\author{Ke Tang}
\cortext[cor]{Email: \{wrui1108@mail.ustc.edu.cn, ketang@ustc.edu.cn\}}
\address{Nature Inspired Computation and Applications Laboratory (NICAL),\\ School of Computer Science and Technology,\\ University of Science and Technology of China, Hefei, Anhui, China, 230027}

\begin{document}

  \begin{abstract}
    Feature selection is an important pre-processing step for many pattern classification tasks. Traditionally, feature selection methods are designed to obtain a feature subset that can lead to high classification accuracy. However, classification accuracy has recently been shown to be an inappropriate performance metric of classification systems in many cases. Instead, the Area Under the receiver operating characteristic Curve (AUC) and its multi-class extension, MAUC, have been proved to be better alternatives. Hence, the target of classification system design is gradually shifting from seeking a system with the maximum classification accuracy to obtaining a system with the maximum AUC/MAUC. Previous investigations have shown that traditional feature selection methods need to be modified to cope with this new objective. These methods most often are restricted to binary classification problems only. In this study, a filter feature selection method, namely MAUC Decomposition based Feature Selection (MDFS), is proposed for multi-class classification problems. To the best of our knowledge, MDFS is the first method specifically designed to select features for building classification systems with maximum MAUC. Extensive empirical results demonstrate the advantage of MDFS over several compared feature selection methods.
  \end{abstract}
  \begin{keyword}
    Feature selection \sep MAUC \sep Filter methods \sep Pattern classification
  \end{keyword}
  \maketitle

  \section{Introduction}
    Feature selection is an important data pre-processing technique in the machine learning and data mining community \cite{Bluma1997, Guyon2003, Hua2009}. By selecting a feature subset from the original feature set, the time and storage requirements of classification tasks are reduced. In addition, reducing the number of features may facilitate data visualization and understanding, or even improve the performance of classification systems \cite{Guyon2003}. Generally speaking, feature selection methods can be divided into two categories, i.e., filter and wrapper \cite{Kohavi1997}. In a filter method, the whole selection procedure is conducted solely based on the data set. On the other hand, a wrapper method employs the classifier that will be used in the classification task afterward to evaluate the merit of each candidate feature subset. It is well known that, in general, filter methods are more computationally efficient, while wrapper methods will lead to better classification performance for the specific classifier \cite{Guyon2003}. In recent years, with the emergence of many large-scale problems that may involve thousands of features (e.g., gene expression \cite{Tang2006, Liu2010} and text classification \cite{Forman2003}), the efficiency of feature selection methods has become of greater concern to both researchers and practitioners. Therefore, filter methods, although sometimes leading to inferior classification performance, are attracting increasing interest.

    In filter methods, since no classifier is used to evaluate candidate feature subsets, alternative metrics are needed to evaluate their utility. Due to the consideration of computational efficiency, many so-called feature ranking methods evaluate the utility of individual features, and pick out the top ones. Fisher's ratio \cite{Pruzansky1964}, Pearson's correlation coefficient \cite{Miyahara2000}, Chi-square \cite{Liu1995}, information gain \cite{Hunt1966, Liu2009}, symmetrical uncertainty \cite{Press1988} and distance discriminant \cite{Liang2008} have all been utilized as metrics for this purpose. In addition to measuring the utility of individual features, the Relief methods \cite{Kira1992, Robnik-Sikonja2003} further take the interaction between features and local characteristics of the sample space into consideration, and thus are more likely to obtain a good feature subset rather than a set of good individual features. However, all these methods may suffer from selecting redundant features that provide no additional information but cause more computation time for classification. To address this disadvantage, more recent filter methods, such as the minimal Redundancy Maximal Relevance (mRMR) \cite{Peng2005} and Fast Correlation Based Filter (FCBF) methods \cite{Yu2004}, are equipped with schemes to exclude redundant features. Since these methods usually involve calculating of the relevance between pairs of features, a major payoff of them is the much higher computational cost.

    A good feature selection method should not only be efficient, but also guarantee high classification performance. Traditionally, this means that the feature selection process should select a feature subset that leads to high classification accuracy. Most state-of-the-art methods, including those mentioned above, have been demonstrated to be effective with regards to this objective. However, recent progresses in the machine learning area and new application domains have revealed that accuracy itself is not necessarily a good performance metric \cite{Huang2005}. First, using accuracy assumes that the prior probabilities of different classes in the data sets are approximately equal, which is not the case in many real-world applications (such as imbalanced learning problems \cite{He2009}). Second, using accuracy assumes that different types of misclassifications induce the same cost, which does not hold in many real-world applications (such as cost sensitive learning problems \cite{Elkan2001}). To address these shortcomings of accuracy, two alternative metrics, called Area Under the receiver operating characteristic Curve (AUC) \cite{Bradley1997, Fawcett2006} and its multi-class extension, named MAUC \cite{Hand2001}, have been introduced in recent years. Specifically, AUC is used to evaluate binary classifiers and MAUC for multi-class ones. They measure the performance of classifiers without making implicit assumptions about the prior probability of classes or the misclassification costs. Therefore, classification systems with maximized AUC/MAUC can be more useful for real-world problems that involve unequal, unknown or even changing class distribution and misclassification costs \cite{Provost2001}. Moreover, it has been theoretically proved that AUC is more powerful than accuracy for discriminating classification systems, and extensive empirical studies showed that similar conclusion also holds for MAUC \cite{Huang2005}. In other words, even for balanced data sets that do not involve different costs, AUC and MAUC are still superior to accuracy in the sense that they facilitate choosing the best classification system from a number of candidates. Therefore, the aim of designing a classification system is now gradually shifting from maximizing the accuracy of the system to maximizing its AUC or MAUC. Hereafter, we refer to classification systems that are designed according to this new objective as AUC/MAUC-oriented classification systems.

    Given the difference between accuracy and AUC/MAUC, it is interesting to ask whether traditional feature selection methods can cope with the new challenge raised by AUC/MAUC-oriented classification systems. Recently, some initial studies have been carried out to address this issue. In \cite{Chen2008}, AUC is employed to rank features directly. This work was then further extended in \cite{Wang2009}. Empirical studies have shown that these two methods, although derived from traditional filter methods with minor modifications, significantly outperform traditional methods. This observation is not unexpected since it has been stated that a successful feature selection method should consider the objective of the classification systems \cite{Tsamardinos2003}. However, both of the above methods focus only on binary classification problems. To the best of our knowledge, no work has been published in literature to address multi-class classification problems. Yet, multi-class problems are very common in practice and there is a need for suitable MAUC-oriented feature selection methods. We therefore propose in the paper a novel feature selection method for MAUC-oriented classification systems.

    The proposed method, namely MAUC Decomposition based Feature Selection (MDFS), is in essence a filter method. In MDFS, a multi-class problem is first divided into a batch of binary class sub-problems in one-versus-one manner (i.e., each sub-problem consists of a pair of classes \cite{Furnkranz2002}). After that, AUC is used to rank all features within each sub-problem. Thus, a feature ranking list can be obtained for each sub-problem. Finally, the sub-problems are accessed iteratively. Every time that a sub-problem is considered, one feature is picked out from the corresponding feature ranking list. In this way, the ``siren pitfall'' phenomenon \cite{Forman2004} that is usually encountered in multi-class feature selection is avoided. Extensive empirical studies have been conducted to compare MDFS to 8 other feature selection methods on 8 multi-class data sets with 4 different types of classifiers. The results clearly demonstrated the superiority of MDFS.

    The rest of the paper is organized as follows: Related feature selection methods are briefly reviewed in Section 2. In Section 3, AUC and MAUC are introduced. After that, MDFS is described in detail in Section 4. Section 5 presents the experimental setup and results. Finally, conclusions and discussions are given in Section 6.
  \section{Feature Selection Methods Revisited}
    In this section, we will review the filter methods that are closely related to our work, including three feature ranking methods, the SpreadFx approach proposed in \cite{Forman2004}, the ReliefF method \cite{Robnik-Sikonja2003}, which is the multi-class extension of the Relief method, and the minimal Redundancy Maximal Relevance (mRMR) method \cite{Peng2005}.

    The main notations used in this paper are summarized as follows. $D=\{x_1, x_2, \ldots, x_n\}$ is the training data set, where $x_i\,(1\leq{i}\leq{n})$ is an instance. $F =\{f_1, f_2, \ldots, f_m\}$ is the original feature set, where $f_i\,(1\leq{i}\leq{m})$ is a feature. In addition, the class variable is denoted by $y$, whose value can be one of $\{1, 2,\ldots, c\}$ for each instance.
  \subsection{Feature Ranking methods}
    Feature ranking methods score each feature individually according a pre-defined criterion. Then the top $K$ (a user-defined number) features with the largest scores will be selected. Methods of this category are very popular due to their high computational efficiency and simplicity. In the following, we will briefly revisit three popular feature ranking methods.
  \subsubsection{Feature Ranking Based on Chi-Square}
    The feature ranking method based on the Chi-square metric \cite{Liu1995} utilizes a very simple selection strategy.  For a nominal feature $f_i$, its Chi-square statistics CHI for the class variable $y$ can be calculated as follows,
    \begin{equation}
      CHI=\sum_{jk}{\frac{(O_{jk}-E_{jk})^2}{E_{jk}}}
    \end{equation}
    where $O_{jk}$ is the number of instances with feature value $f_{i}=f_{ij}$ ($f_{ij}$ is a possible value of feature $f_i$) and $y=k$.
    \begin{equation}
      E_{jk}=\frac{O_{j\cdot}\times{O_{\cdot{k}}}}{n}
    \end{equation}
    where $O_{j\cdot}$ is the number of instances with feature value $f_{i}=f_{ij}$, and $O_{\cdot{k}}$ is the number of instances with $y=k$.
  \subsubsection{Feature Ranking Based on Symmetrical Uncertainty}
    Instead of the Chi-square statistics, this method use the symmetrical uncertainty \cite{Press1988} statistics SU to rank features. As a variant of mutual information, symmetrical uncertainty avoids the bias of mutual information to features with many distinct values and lies in the range [0,1]. For a nominal feature $f_{i}$ and the class variable $y$, SU between them is calculated as,
    \begin{equation}
      SU=\frac{2\times{I(f_i; y)}}{H(f_{i})+H(y)}
    \end{equation}
    where
    \begin{gather}
      H(f_{i})=-\sum_{j}{p(f_i=f_{ij})\log{p(f_i=f_{ij})}}\\
      H(y)=-\sum_{k=1}^{c}{p(y=k)\log{p(y=k)}}
    \end{gather}
    are the entropy of $f_{i}$ and $y$ respectively, and
    \begin{equation}
     \label{MI}
      I(f_i; y)=H(y)-H(y|f_{i})
    \end{equation}
    denotes the mutual information between feature $f_{i}$ and class variable $y$. Furthermore,
    \begin{equation}
      H(y|f_{i})=-\sum_{jk}{p(f_i=f_{ij},y=k)\log{p(y=k|f_i=f_{ij})}}
    \end{equation}
    is the conditional entropy of $y$ given $f_{i}$.
  \subsubsection{Feature Ranking Based on Distance Discriminant}
    To select features that can separate different classes while keep the instances in the same class close to one another, Feature Selection based on Distance Discriminant (FSDD), was proposed in \cite{Liang2008}. FSDD calculates the utility of feature $f_i$ according to Eq. (\ref{fsdd}),
    \begin{equation}
        \label{fsdd}
      \frac{1}{\sigma_{i}^{2}}\bigg[\sigma_{i}^{''2}-\beta{\sum_{k=1}^{c}{\frac{n_{k}}{n}\sigma_{i}^{'2}(k)}}\bigg]
    \end{equation}
    where $\beta$ is a tuning parameter (by default, $\beta=2$), $n_{k}$ is the number of instances in the $k$-th class.
    \begin{equation}
      \sigma_{i}^{2}=\frac{1}{n-1}\sum_{x_{j}\in{D}}{(x_{j}^{i}-\bar{f}_i)^2},\,\,\,\,\,\,\bar{f}_{i}=\frac{1}{n}\sum_{x_j\in{D}}{x_{j}^{i}}
    \end{equation}
    are the variance and mean of feature $f_i$ over all instances in data set $D$. Here, $x_{j}^{i}$ is the value of feature $f_i$ for instance $x_j$.
    \begin{equation}
      \sigma_{i}^{'2}(k)=\frac{1}{n_{k}-1}\sum_{x_{j}\in{class(k)}}{\Big(x_{j}^{i}-\bar{f}_{i}(k)\Big)^2},\,\,\,\,\,\,\bar{f}_{i}(k)=\frac{1}{n_{k}}\sum_{x_{j}\in{class(k)}}{x_{j}^{i}}
    \end{equation}
  are the variance and mean of feature $f_i$ over all instances that belong to the $k$-th class.
    \begin{equation}
      \sigma_{i}^{''2}=\sum_{k=1}^{c}{\frac{n_{k}}{n}\Big(\bar{f}_{i}(k)-\bar{f}_{i}\Big)^2}
    \end{equation}
    is the weighted variance of the feature $f_i$ over $c$ differen classes.
  \subsection{SpreadFx}
    To overcome the ``siren pitfall'' phenomenon that adheres to traditional feature ranking methods in multi-class problems (we will detail this issue in Section 4.1), the SpreadFx \cite{Forman2004} feature selection approach was proposed. Generally speaking, SpeadFx methods first decompose the multi-class problem into $c$ binary sub-problems in one-versus-all manner (i.e., each sub-problem consists of one class as positive class and all the other classes jointly form a negative class). Then, a feature ranking list will be obtained on each sub-problem. Finally, features are selected by applying some dynamic scheduling policy to the ranking lists. Two key components need to be specified when employing a SpreadFx type method: the feature ranking method and the the dynamic scheduling policy. In practice, the former can be any feature ranking method (for example, any of the three methods described above). As for the latter, it has been shown in \cite{Forman2004} that selecting sub-problem one by one in turn (the Round-Robin policy) performed satisfactorily.
  \subsection{ReliefF}
     The Relief methods first calculate a weight for each feature. Then they select the features with the largest weights \cite{Kira1992}. Different from feature ranking methods, the weights of features are calculated in an iterative way. The calculation is based on the assumption that instances belonging to the same class and close to one another should have similar values on a useful feature, while instances that are close to one another but are from different classes should have quite different values. At each iteration, Relief methods choose one instance and its nearest neighbors in each class. Then, the difference between this instance and its neighbors on every feature are employed to update the weights of features. This procedure is repeated for a pre-defined number of iterations. As an extension of the original Relief method, ReliefF was proposed to tackle multi-class problems and is more robust to noisy and missing values in data sets \cite{Robnik-Sikonja2003}.
  \subsection{Minimal Redundancy Maximal Relevance Method}
    Some recent filter feature selection methods are equipped with schemes to explicitly exclude redundant features. A representative method of them is the minimal Redundancy Maximal Relevance (mRMR) method \cite{Peng2005}. Specifically, mRMR first evaluates the \emph{relevance} of each feature based on its mutual information (Eq. (\ref{MI})) with the class variable, then the feature with largest relevance score is selected in the first iteration. After that, features are selected one at a time according to the following criterion:
    \begin{equation}
      f_{selected}=\mathop{\arg\max}_{f_i\in{F-S}}{\bigg[I(f_i, y)-\frac{1}{|S|}\sum_{f_j\in{S}}{I(f_i; f_j)}\bigg]}
    \end{equation}
    where $f_{selected}$ denotes the feature being selected at each iteration, $S$ is the currently selected feature subset, and $|S|$ is its cardinality. The algorithm terminates when a pre-defined number of features have been selected. Since mRMR detects the redundancy among features by calculating the mutual information between pairs of features, it has a higher computational complexity in comparison to other filter methods.
  \section{Area Under the ROC Curve and its Multi-class Extension}
  \subsection{Area Under the ROC Curve}
    Assume that a data set consists of $n$ instances, with $P$ of them belonging to class 1 (positive class) and $N$ of them belonging to class 2 (negative class). AUC can be used to evaluate the performance of a classification system on this binary class data set. Generally speaking, most mainstream classifiers allow assigning a numerical score (e.g., probability of the instance belonging to the positive class) to each of the $n$ instances after training. Then, the AUC value of this classifier can be calculated based on these  $n$ scores and the corresponding true class labels. To be specific, the AUC value of a classifier equals to the probability that a randomly chosen positive instance will be assigned a larger score than a randomly chosen negative instance \cite{Hanley1982}. As  an example, given a classifier whose AUC value is 0.85 on data set $D$, for a randomly chosen positive instance $x_1$ and a randomly chosen negative instance $x_2$ from $D$, the expected probability that $x_1$ will get higher score than $x_2$ is 0.85.

    Let a classifier $h(x_i)\rightarrow\mathbb{R}$ outputs a numerical score to indicate the confidence that $x_i$ belongs to the positive class. AUC can be calculated as,
    \begin{equation}
      \label{auc}
      AUC=\frac{\sum_{x_i\in{class(1)};\,x_j\in{class(2)}}{s(x_i,x_j)}}{P\times{N}}
    \end{equation}
    where $s(x_i,x_j)$ is defined as:
    \begin{equation}
      s(x_i,x_j)=
      \begin{cases}
        1 & \text{if  } h(x_i)>h(x_j);\\
        0.5 & \text{if  } h(x_i)=h(x_j);\\
        0 & \text{if  } h(x_i)<h(x_j).
      \end{cases}
    \end{equation}
  \subsection{MAUC}
    Although AUC has been studied extensively in the literature, it is only applicable to binary classification problems. A simple extension of AUC which can be used in multi-class problems was proposed in \cite{Hand2001}. For a multi-class classification problem containing $c$ classes, a classifier assigns $c$ scores to every instance. Each score corresponds to one of the $c$ classes and indicates the confidence that the instance belongs to this class. Hence, the scores for all the $n$ instances can be represented by an $n$-by-$c$ matrix, the columns and rows of which correspond to classes and instances respectively. Let $A_{ij} \,(A_{ji})$ be the AUC calculated according to the $i$-th ($j$-th) column of the matrix with respect to the instances from class $i$ and class $j$, MAUC is defined as follows,
    \begin{equation}
      \label{mauc}
      MAUC=\frac{2}{c\times(c-1)}\sum_{i=1}^{c}{\sum_{j=i+1}^{c}{\frac{A_{ij}+A_{ji}}{2}}}.
    \end{equation}

    From Eq. (\ref{mauc}), we can see that the MAUC value of a classification system is actually the average AUC value over all $\frac{c(c-1)}{2}$ one-versus-one binary sub-problems. In other words, this means that maximizing the AUC value of each binary sub-problem is of equal importance in calculating a classification system's MAUC.
  \section{Feature Selection for MAUC-Oriented Classification Systems}
    Given the difference between accuracy and MAUC, the question that is addressed in this study is: How can the feature selection process facilitates the construction of MAUC-oriented classification systems? In general, three methodologies can be employed for this purpose. First, traditional feature selection methods can be applied directly.  Second, in analogy to \cite{Chen2008}, MAUC can be used as a relevance metric to rank features. Third, one may also develop a novel feature selection method. We will start by considering the efficacy of the former two methodologies. Then, the MDFS method, which belongs to the third methodology, will be described in detail.
  \subsection{Using Traditional Feature Selection Methods Directly}
    As mentioned before, previous work showed that traditional feature selection methods may be easily outperformed by new methods specifically designed for AUC-oriented classification systems on binary problems. It is natural to expect that such a situation also holds for multi-class problems, although solid evidence is absent in the literature. In fact, traditional feature selection methods might be unsuitable for MAUC-oriented classification systems not only because of the difference between accuracy and MAUC, but also due to the so-called ``siren pitfall'' phenomenon that can occur in traditional feature ranking methods: Since the difficulties of separating different classes are usually different in multi-class problems, features that perform well on easy sub-problems (i.e., the sub-problems consists of classes that are easy to separate) usually can obtain a relatively higher utility scores, while features that perform well on difficult sub-problems usually obtain lower scores. As a result, when these features are compared to each other, those which perform well on easy sub-problems are more likely to be selected \cite{Forman2004}. In consequence, easy sub-problems will be focused on more than difficult sub-problems. However, when calculating the MAUC value of a classification system, it is equally important to maximize the AUC value of every sub-problem. Hence, the ``siren pitfall'' phenomenon of feature selection methods may also degenerate the performance of MAUC-oriented classification systems.

    In the literature, the ``siren pitfall'' phenomenon was only reported for text classification systems whose performance were measured by Precision, F-measure, and Recall \cite{Forman2004}. To verify whether it also exists in MAUC-oriented classification systems, we carried out a case study on the Indiana data set \cite{Landgrebe1992}. This hyperspectral imagery data set is a section of scenes taken over northwest Indiana's Indian Pines by the airborne visible/infrared imaging spectrometer (AVIRIS), and consists of 9 classes and 220 features. Similar to the experimental setup in \cite{Forman2004}, the Naive Bayes classifier was applied to every one-versus-one sub-problem using all the features. Two feature selection methods, namely feature ranking based on the Chi-square metric (CHI) and feature ranking based on the symmetrical uncertainty metric (SU), were employed to select 100 features on the whole data set (global best features). Then, they were applied to each sub-problem separately to rank all the 220 features. The higher the rank (let rank 1 be the highest rank), the more useful is this feature for the corresponding sub-problem. If the global best features work well on a sub-problem, they will get high rank on it. Fig. 1 presents the AUC obtained by the Naive Bayes classifier on the 36 sub-problems. Figs. 2 and 3 illustrate the ranks of the global best 100 features on each sub-problem. It can be observed that the global best features selected by both CHI and SU got rather low ranks on those difficult sub-problems (i.e., the sub-problems corresponding to smaller AUC in Fig. 1). For example, for the 34th sub-problem, most of the 100 global best features were not within the top features. This observation suggests that the ``siren pitfall'' should also be taken care of for MAUC-oriented systems. In \cite{Forman2004}, SpreadFx was proposed to cope with the ``siren pitfall'' phenomenon. However, it was not designed for MAUC-oriented classification systems and may not suit the aim of MAUC maximization well. Hence, new feature selection methods need to be developed.
    \begin{figure}[!htbp]
      \centering
      \includegraphics[width=0.75\textwidth]{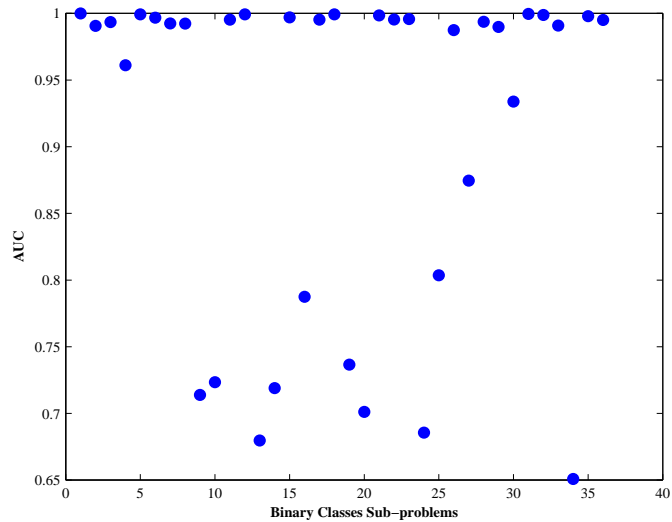}
      \caption{Difficulty of the binary classification sub-problems of the Indiana data set. The larger the AUC, the easier the corresponding sub-problem is. }
    \end{figure}
    \begin{figure}[!htbp]
      \centering
      \includegraphics[width=0.75\textwidth]{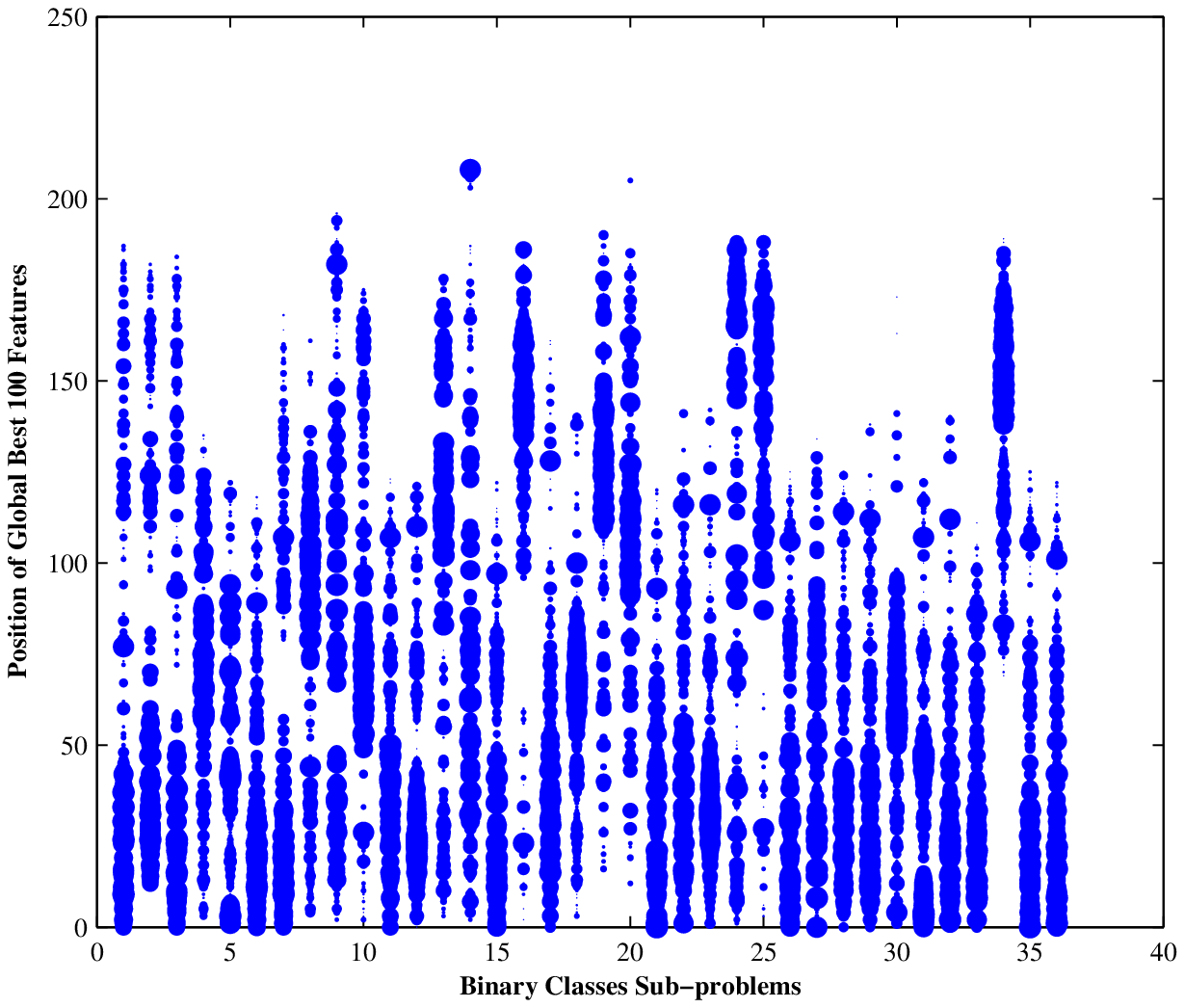}
      \caption{Ranks of the top 100 global best features on the 36 sub-problems. The 100 features are selected by CHI method.}
    \end{figure}
    \begin{figure}[!htbp]
      \centering
      \includegraphics[width=0.75\textwidth]{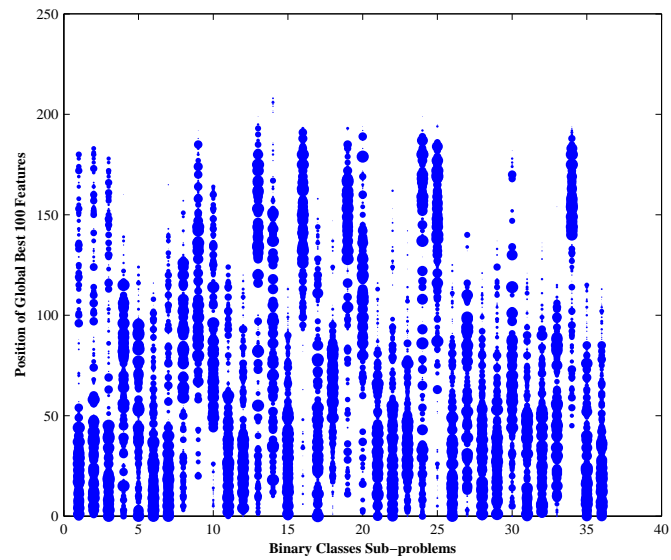}
      \caption{Ranks of the top 100 global best features on the 36 sub-problems. The 100 features are selected by SU method.}
    \end{figure}
  \subsection{Using MAUC as Relevance Metric Directly}
    Since the value of  a feature on each instance can be interpreted as the output of a single feature classifier, a straightforward feature selection method for MAUC-oriented classification systems is using MAUC as the relevance metric to rank features directly (referred to as MAUCD hereafter). The MAUC score of a feature can be calculated in two steps: First, decompose the multi-class problem to a batch of one-versus-one binary class sub-problems, and then calculate the AUC score of this feature on every sub-problem. Second, calculate this feature's MAUC score using Eq. (\ref{mauc}). In this method, the utility of a feature is measured by averaging its utility over all sub-problems, and thus quite a few features that favor easy sub-problems will get large MAUC scores. Consequently, these features are more likely to be selected than those features that are more useful for difficult sub-problems. In other words, directly using MAUC to rank and select features can induce the ``siren pitfall'' phenomenon as well, and thus might not be an ideal solution for MAUC-oriented classification systems. Furthermore, it is likely that different features may be useful for different sub-problems, and we can anticipate that conducting feature selection on every sub-problem separately and collecting all the obtained feature subsets will form a feature subset that yields good performance. Therefore, instead of the direct use of MAUC as feature ranking metric, we design the MAUC Decomposition based Feature Selection (MDFS) method.
  \subsection{MAUC Decomposition based Feature Selection Method}
    Give a data set $D$, each instance $x_i$ may belong to one of $c\,(c>2)$ classes and is represented by $m$ features. The MDFS method works as follows. First, $D$ will be decomposed into $\frac{c(c-1)}{2}$ binary class sub-problems in one-versus-one manner (i.e., each sub-problem consists of a pair of classes). Then, the features are ranked according to their AUC scores on every sub-problem. This leads to $\frac{c(c-1)}{2}$ feature ranking lists. After that, feature selection is carried out iteratively. In each iteration, a sub-problem is randomly chosen and the previously unselected feature with the highest rank in the corresponding feature ranking list is moved to the selected feature subset. Since the AUC score is used to rank features in every sub-problem, MDFS can only deal with numerical, ordered and binary type features. Nominal features which take more than two possible values need to be converted to appropriate numerical features before using MDFS. Algorithm 1 presents the main steps of MDFS.
    \begin{algorithm}[htbp]
    \caption{The MDFS Feature Selection Algorithm}
    \LinesNumbered
    \SetKwInOut{Input}{Input}
    \SetKwInOut{Output}{Output}
    \Input{Multi-class data set $D$, a user set feature subset size $K$}
    \Output{A set $S$ contains selected features}
    \BlankLine
    $S\leftarrow\emptyset$\;
    \For{$i\leftarrow{1}$ \KwTo $c$}{
        \For{$j\leftarrow{i+1}$ \KwTo $c$}{
            get the binary sub-problem $D_{ij}$ in one-versus-one manner\;
            get the feature rank list $L_{ij}$ according to their AUC scores on $D_{ij}$;
        }
    }
    \While{$|S|<K$}{
        select a sub-problem $D_{ij}$ randomly\;
        identify the best feature $f$ from $L_{ij}$\;
        pop $f$ out of $L_{ij}$\;
        \If{ $f\notin{S}$}{
            put $f$ into $S$\;
        }
    }
    \KwRet{$S$}
    \end{algorithm}
  \subsection{Computational Complexity of MDFS}
    In this subsection, the time complexity of MDFS is analyzed and compared to some other existing feature selection methods. Let the number of instances in the $i$-th class of data set $D$ be $n_i$. The complexity of calculating the AUC score of one feature on the sub-problem consisting of the $i$-th and $j$-th class then is $O\Big((n_i+n_j)\log(n_i+n_j)\Big)$ \cite{Hand2001}. Since MDFS requires calculating the AUC scores of $m$ features on every sub-problem, its time complexity is:
    \begin{equation}
      O\Big(m\sum_{i=1}^{c}{\sum_{j=i+1}^{c}{(n_i+n_j)\log(n_i+n_j)}}\Big)
    \end{equation}
    Since
    \begin{equation}
      m\sum_{i=1}^{c}{\sum_{j=i+1}^{c}{(n_i+n_j)\log(n_i+n_j)}}\\
      \leq{m\sum_{i=1}^{c}{\sum_{j=i+1}^{c}{n\log{n}}}}\\
      =\frac{c(c-1)}{2}{m}{n\log{n}}
    \end{equation}
   and the class number $c$ is usually small in practice, the complexity of MDFS is roughly $O(mn\log{n})$. The main computational cost of MAUCD is also induced by calculating the AUC scores of features on all binary sub-problems. Hence, the complexity of MAUCD is the same as that of MDFS. According to \cite{Liang2008}, the time complexity of FSDD is $O(mn)$. The CHI and SU methods are designed to deal with nominal features. For numerical features, a discretization procedure is needed to convert the numerical features into nominal ones. A typical discretization method requires sorting the numerical values first, and then scanning over the sorted values to convert an interval of continuous values to a discrete value \cite{Liu2002}. Hence, the complexity of these two feature ranking methods in dealing with data sets consists of numerical type features is also $O(mn\log{n})$. Following this analysis, the complexity of SpreadFx feature selection approach using one of these feature ranking methods to rank features on every sub-problem is $O(cmn\log{n})$. Again, omitting the constant $c$, SpreadFx's complexity is $O(mn\log{n})$ as well. The complexity of ReliefF is $O(tmn)$, where $t$ is the number of iterations to update features' weights \cite{Robnik-Sikonja2003}. The configuration of $t$ involves many factors and is a non-trivial task \cite{Robnik-Sikonja2003, Liu2004}. If $t$ is set to $\log{n}$ as recommended in \cite{Hong1997}, the complexity of ReliefF will also be $O(mn\log{n})$. In addition, since mRMR detects the redundancy among features by calculating the mutual information between pairs of features, the time complexity of mRMR is $O(m^2n\log{n})$. To summarize, the computational complexity of MDFS is comparable to that of existing filter feature selection methods.
  \section{Experiments}
    Experimental studies have been carried out to evaluate the performance of MDFS. Our experiments were designed based on three considerations. First, the efficacy of MDFS needs to be verified against traditional filter feature selection methods. Second, since the focus of this work is filter methods, the experiments should not be restricted to a specific type of classifier. Finally, the experiments should be conducted on data sets with sufficient numbers of features. Otherwise, it is not necessary to conduct feature selection at all. Having these considerations in mind, 9 filter methods (including MDFS) and 4 different types of classifiers have been selected for the comparison. Altogether 36 combinations of feature selection methods and classifiers have been applied to 8 multi-class data sets collected from various domains with more than 60 features.
  \subsection{Data Sets}
     Eight benchmark multi-class data sets from various domains were collected for our experiments. The ISOLET data set \cite{Asuncion2007}, MNIST data set \cite{Asuncion2007}, USPS data set \cite{Hull1994} are handwriting recognition problems. The Phoneme data set \cite{Hastie2001} is from the speech recognition field. The Washington data set \cite{Neher2005} and Indiana data set \cite{Landgrebe1992} are hyperspectral imagery data sets. More details about these two data sets can be found in \cite{Zhu2010}. Synthetic data set \cite{Asuncion2007} is a synthetically generated control chart data set and Thyroid \cite{Yukinawa2006} is a microarray data set. Table 1 summarizes the information about these data sets.
    \begin{table}
      \caption{Summary of 8 benchmark data sets}
      \centering
      \begin{tabular}{lccc}
        \toprule
        Data Set& No. of Instances & No. of Features & No. of Classes\\
        \midrule
        ISOLET & 7797 & 617 & 26\\
        MNIST  & 10000& 784 & 10\\
        USPS   & 7291 & 256 & 10\\
        Phoneme& 4509 & 256 & 5 \\
        Washington & 11200 & 210 & 7\\
        Indiana & 9345 & 220 & 9\\
        Synthetic & 600 & 60 & 6\\
        Thyroid & 168 & 2000 & 4\\
        \bottomrule
      \end{tabular}
    \end{table}
  \subsection{Experimental Setup}
    The three feature ranking methods introduced in Section 2.1, and ReliefF were picked as the baseline feature selection methods in our experiments. To compare with SpreadFx, we set CHI and SU separately as the feature ranking method in SpreadFx, and employed \emph{Round-Robin} as the scheduling scheme. The resulting algorithms are referred to as SpreadFx [Round-Robin, CHI] (SCHI) and SpreadFx [Round-Robin, SU] (SSU) respectively. Besides, the mRMR feature selection method introduced in Section 2.4 and MAUCD introduced in Section 4.2 were also included for comparison.

    Since none of of these methods can automatically decide how many features should be selected in a given problem, we compared them on different feature subset sizes from 10 to the 100 with an interval of 10. Since the Synthetic data set only consists of 60 features, 6 feature subset sizes were considered on this data set.

    In order to examine whether MDFS is biased towards a certain type of classifier, 4 different types of classifiers were used in our experiments: 1-Nearest Neighbor (1NN) \cite{Aha1991}, C4.5 \cite{Quinlan1993}, Naive Bayes \cite{John1995}, and SVM with RBF kernel function \cite{Corinna1995}.

    All the compared feature selection methods and classifiers were implemented on the WEKA platform \cite{Witten2000}, the number of iterations $t$ used to update features' weight in ReliefF method was set to $\log{n}$, where $n$ is the number of training instances. The parameters $c$ and $\gamma$ of SVM were set to the values which maximize the average MAUC in a 3-fold cross-validation on the training data set of a 10-fold cross-validation, where $c$ was sampled at $2^{-5}, 2^{-3}, \ldots, 2^{15}$, and $\gamma$ was sampled at $2^{-15}, 2^{-13}, \ldots, 2^3$. All other parameters and implementations followed the default configuration of WEKA.

    All classification systems with different configurations (classification algorithm, feature selection method, feature subset size) were evaluated on the 8 data sets by applying 10-fold cross-validation for 10 times. The average MAUC values of the classification systems with the same feature subset size and classification algorithm were used as the indicator to compare different feature selection methods. The Wilcoxon signed-rank test with $95\%$ confidence level was employed to examine whether the differences between the performance of MDFS and other feature selection methods are statistically significant.
  \subsection{Results}
    The results of our experiments are summarized in Table 2 to Table 9 and Fig. 4. Tables 2 to 9 present the MAUC value for each configuration, one table for one data set. The results of the pairwise Wilcoxon signed-rank test between MDFS and 8 other feature selection methods are also labeled as superscript on the MAUC values of corresponding classification systems. $\dag$ ($\ddag$) means the corresponding result is statistically worse (better) than the result of MDFS. Otherwise, no difference has been detected by the Wilcoxon test. The largest MAUC value of each configuration is in boldface.

    For the ISOLET data set, MDFS outperformed all the other 8 compared methods on every classifier when the feature subset size is larger than 10. For the MNIST data set, except for two configurations (SVM + feature subset size 90, 100), MDFS always obtained the best results. A similar situation can be observed with the USPS data set. Hence, the superiority of MDFS has been proved on these three data sets. On the speech recognition data set (i.e., the Phoneme data set), none of the feature selection methods dominated the others. On the Washington data set, MDFS outperformed others when cooperated with Naive Bayes and 1NN. For other configurations, SSU or SCHI performed better. For Indiana data set, mRMR worked significantly well with Naive Bayes classifier, and SSU or SCHI performed better otherwise. For the synthetically control chart data set, classifiers working with MDFS resulted in the largest MAUC in most cases. For the Thyroid data set, MDFS performed well with 1NN and SVM. As for Naive Bayes and C4.5, MDFS is comparable with the other methods.

    In general, we can see that the classification systems employing MDFS as feature selection method have led to the largest MAUC value more often than not. In the cases that MDFS was not the best, it still outperformed some compared methods. Specifically, MAUCD was almost always inferior to MDFS, this is not surprising due to the reasons stated in Section 4.2. The three feature ranking methods, CHI, SU and FSDD were clearly outperformed by MDFS throughout our experiments, which also coincided with our analysis in Section 4.1. Despite its appealing performance in accuracy-oriented classification systems, the mRMR method was dominate by MDFS in MAUC-oriented classification systems in most of the cases throughout our experiments. SCHI and SSU were supposed to be the biggest challenger to MDFS. However, the overall results on these 8 data sets clearly demonstrated the advantage of MDFS.

    Fig. 4 summarizes the results of the Wilcoxon signed rank tests conducted between MDFS and the other 8 compared methods. There are 4 sub-graphs, corresponding to the comparisons on 4 different types of classifiers. The hight of each bar is the number of times that MDFS win (draw or lose) the counterpart feature selection method on the corresponding classifier over all feature subset sizes and data sets. It can be clearly seen that MDFS performed significantly better than all the compared methods on all the 4 classifiers.

    Finally, Table 10 shows the runtime of every feature selection method on the 8 data sets. As analyzed in Section 4.4, the computational complexity of MDFS is comparable with that of most of the compared feature selection methods (except the FSDD method and the mRMR method). However, due to the constant factor (e.g., number of classes $c$) in the complexity analysis and implementation details of these algorithms, the actual runtime may deviate from the complexity analysis and is only indicative.
\begin{table}
\centering
\caption{Average MAUC obtained with the nine compared methods on the ISOLET data set. The results were obtained by repeating 10-fold cross-validation procedure for 10 times. For each classifier and feature subset size, the Wilcoxon signed-rank test with 95\% confidence level is employed to compare MDFS and 8 other methods. The methods that performed significantly worse (better) than MDFS are highlighted with \dag (\ddag). No superscript is used if no statistical significant difference is detected. The largest MAUC value of each configuration is in boldface.}
\newsavebox{\ISOLET}
\begin{lrbox}{\ISOLET}
\begin{tabular}{lr*{10}{c}}
\toprule
& & \multicolumn{10}{c}{Feature Subset Size}\\
\cmidrule(r){3-12}
& & 10 & 20 & 30 & 40 & 50 & 60 & 70 & 80 & 90 & 100\\
\midrule
\multirow{9}{*}{Naive Bayes}
& MDFS & 0.958 & \textbf{0.977} & \textbf{0.982} & \textbf{0.984} & \textbf{0.985} & \textbf{0.985} & \textbf{0.986} & \textbf{0.986} & \textbf{0.986} & \textbf{0.986}\\
& MAUCD & $0.906^{\dag}$ & $0.924^{\dag}$ & $0.917^{\dag}$ & $0.930^{\dag}$ & $0.941^{\dag}$ & $0.948^{\dag}$ & $0.952^{\dag}$ & $0.962^{\dag}$ & $0.973^{\dag}$ & $0.977^{\dag}$\\
& mRMR & $\textbf{0.962}^{\ddag}$ & $0.973^{\dag}$ & $0.977^{\dag}$ & $0.979^{\dag}$ & $0.981^{\dag}$ & $0.982^{\dag}$ & $0.983^{\dag}$ & $0.984^{\dag}$ & $0.984^{\dag}$ & $0.985^{\dag}$\\
& FSDD & $0.941^{\dag}$ & $0.955^{\dag}$ & $0.961^{\dag}$ & $0.962^{\dag}$ & $0.966^{\dag}$ & $0.972^{\dag}$ & $0.974^{\dag}$ & $0.977^{\dag}$ & $0.980^{\dag}$ & $0.980^{\dag}$\\
& SCHI & $0.961^{\ddag}$ & $0.976$ & $0.978^{\dag}$ & $0.978^{\dag}$ & $0.978^{\dag}$ & $0.978^{\dag}$ & $0.978^{\dag}$ & $0.978^{\dag}$ & $0.978^{\dag}$ & $0.979^{\dag}$\\
& SSU & $0.954^{\dag}$ & $0.974^{\dag}$ & $0.975^{\dag}$ & $0.977^{\dag}$ & $0.976^{\dag}$ & $0.975^{\dag}$ & $0.976^{\dag}$ & $0.975^{\dag}$ & $0.976^{\dag}$ & $0.975^{\dag}$\\
& CHI & $0.945^{\dag}$ & $0.954^{\dag}$ & $0.969^{\dag}$ & $0.972^{\dag}$ & $0.972^{\dag}$ & $0.973^{\dag}$ & $0.974^{\dag}$ & $0.976^{\dag}$ & $0.975^{\dag}$ & $0.975^{\dag}$\\
& SU & $0.904^{\dag}$ & $0.928^{\dag}$ & $0.946^{\dag}$ & $0.957^{\dag}$ & $0.961^{\dag}$ & $0.971^{\dag}$ & $0.974^{\dag}$ & $0.976^{\dag}$ & $0.978^{\dag}$ & $0.978^{\dag}$\\
& ReliefF & $0.911^{\dag}$ & $0.943^{\dag}$ & $0.956^{\dag}$ & $0.963^{\dag}$ & $0.969^{\dag}$ & $0.973^{\dag}$ & $0.976^{\dag}$ & $0.979^{\dag}$ & $0.980^{\dag}$ & $0.982^{\dag}$\\
\midrule
\multirow{9}{*}{C4.5}
& MDFS & 0.836 & \textbf{0.886} & \textbf{0.904} & \textbf{0.911} & \textbf{0.917} & \textbf{0.921} & \textbf{0.923} & \textbf{0.924} & \textbf{0.927} & \textbf{0.928}\\
& MAUCD & $0.727^{\dag}$ & $0.761^{\dag}$ & $0.768^{\dag}$ & $0.779^{\dag}$ & $0.802^{\dag}$ & $0.811^{\dag}$ & $0.813^{\dag}$ & $0.840^{\dag}$ & $0.872^{\dag}$ & $0.876^{\dag}$\\
& mRMR & $0.840$ & $0.869^{\dag}$ & $0.880^{\dag}$ & $0.884^{\dag}$ & $0.890^{\dag}$ & $0.896^{\dag}$ & $0.899^{\dag}$ & $0.902^{\dag}$ & $0.902^{\dag}$ & $0.904^{\dag}$\\
& FSDD & $0.804^{\dag}$ & $0.842^{\dag}$ & $0.867^{\dag}$ & $0.870^{\dag}$ & $0.887^{\dag}$ & $0.900^{\dag}$ & $0.901^{\dag}$ & $0.903^{\dag}$ & $0.909^{\dag}$ & $0.910^{\dag}$\\
& SCHI & $0.862^{\ddag}$ & $0.879^{\dag}$ & $0.896^{\dag}$ & $0.900^{\dag}$ & $0.905^{\dag}$ & $0.909^{\dag}$ & $0.915^{\dag}$ & $0.918^{\dag}$ & $0.922^{\dag}$ & $0.922^{\dag}$\\
& SSU & $\textbf{0.869}^{\ddag}$ & $0.885$ & $0.896^{\dag}$ & $0.899^{\dag}$ & $0.904^{\dag}$ & $0.905^{\dag}$ & $0.910^{\dag}$ & $0.911^{\dag}$ & $0.915^{\dag}$ & $0.917^{\dag}$\\
& CHI & $0.849^{\ddag}$ & $0.853^{\dag}$ & $0.881^{\dag}$ & $0.898^{\dag}$ & $0.906^{\dag}$ & $0.909^{\dag}$ & $0.912^{\dag}$ & $0.915^{\dag}$ & $0.916^{\dag}$ & $0.917^{\dag}$\\
& SU & $0.793^{\dag}$ & $0.816^{\dag}$ & $0.840^{\dag}$ & $0.856^{\dag}$ & $0.861^{\dag}$ & $0.894^{\dag}$ & $0.899^{\dag}$ & $0.905^{\dag}$ & $0.909^{\dag}$ & $0.909^{\dag}$\\
& ReliefF & $0.764^{\dag}$ & $0.819^{\dag}$ & $0.848^{\dag}$ & $0.865^{\dag}$ & $0.878^{\dag}$ & $0.886^{\dag}$ & $0.892^{\dag}$ & $0.899^{\dag}$ & $0.903^{\dag}$ & $0.907^{\dag}$\\
\midrule
\multirow{9}{*}{1NN}
& MDFS & 0.780 & \textbf{0.869} & \textbf{0.903} & \textbf{0.919} & \textbf{0.930} & \textbf{0.937} & \textbf{0.943} & \textbf{0.947} & \textbf{0.951} & \textbf{0.954}\\
& MAUCD & $0.674^{\dag}$ & $0.732^{\dag}$ & $0.746^{\dag}$ & $0.767^{\dag}$ & $0.791^{\dag}$ & $0.802^{\dag}$ & $0.813^{\dag}$ & $0.843^{\dag}$ & $0.869^{\dag}$ & $0.878^{\dag}$\\
& mRMR & $0.774^{\dag}$ & $0.852^{\dag}$ & $0.879^{\dag}$ & $0.889^{\dag}$ & $0.899^{\dag}$ & $0.909^{\dag}$ & $0.911^{\dag}$ & $0.915^{\dag}$ & $0.915^{\dag}$ & $0.917^{\dag}$\\
& FSDD & $0.710^{\dag}$ & $0.795^{\dag}$ & $0.827^{\dag}$ & $0.838^{\dag}$ & $0.864^{\dag}$ & $0.889^{\dag}$ & $0.891^{\dag}$ & $0.902^{\dag}$ & $0.910^{\dag}$ & $0.911^{\dag}$\\
& SCHI & $\textbf{0.788}^{\ddag}$ & $0.852^{\dag}$ & $0.881^{\dag}$ & $0.894^{\dag}$ & $0.906^{\dag}$ & $0.912^{\dag}$ & $0.919^{\dag}$ & $0.927^{\dag}$ & $0.930^{\dag}$ & $0.934^{\dag}$\\
& SSU & $0.780$ & $0.849^{\dag}$ & $0.873^{\dag}$ & $0.885^{\dag}$ & $0.895^{\dag}$ & $0.902^{\dag}$ & $0.910^{\dag}$ & $0.914^{\dag}$ & $0.922^{\dag}$ & $0.926^{\dag}$\\
& CHI & $0.745^{\dag}$ & $0.795^{\dag}$ & $0.862^{\dag}$ & $0.886^{\dag}$ & $0.895^{\dag}$ & $0.906^{\dag}$ & $0.915^{\dag}$ & $0.921^{\dag}$ & $0.925^{\dag}$ & $0.926^{\dag}$\\
& SU & $0.690^{\dag}$ & $0.761^{\dag}$ & $0.800^{\dag}$ & $0.828^{\dag}$ & $0.842^{\dag}$ & $0.880^{\dag}$ & $0.890^{\dag}$ & $0.902^{\dag}$ & $0.913^{\dag}$ & $0.916^{\dag}$\\
& ReliefF & $0.687^{\dag}$ & $0.767^{\dag}$ & $0.810^{\dag}$ & $0.835^{\dag}$ & $0.857^{\dag}$ & $0.869^{\dag}$ & $0.879^{\dag}$ & $0.888^{\dag}$ & $0.895^{\dag}$ & $0.902^{\dag}$\\
\midrule
\multirow{9}{*}{SVM}
& MDFS & 0.808 & \textbf{0.892} & \textbf{0.927} & \textbf{0.944} & \textbf{0.953} & \textbf{0.960} & \textbf{0.965} & \textbf{0.969} & \textbf{0.972} & \textbf{0.974}\\
& MAUCD & $0.707^{\dag}$ & $0.763^{\dag}$ & $0.773^{\dag}$ & $0.801^{\dag}$ & $0.834^{\dag}$ & $0.845^{\dag}$ & $0.852^{\dag}$ & $0.882^{\dag}$ & $0.911^{\dag}$ & $0.919^{\dag}$\\
& mRMR & $0.809$ & $0.871^{\dag}$ & $0.901^{\dag}$ & $0.914^{\dag}$ & $0.929^{\dag}$ & $0.939^{\dag}$ & $0.942^{\dag}$ & $0.947^{\dag}$ & $0.949^{\dag}$ & $0.952^{\dag}$\\
& FSDD & $0.750^{\dag}$ & $0.826^{\dag}$ & $0.859^{\dag}$ & $0.871^{\dag}$ & $0.891^{\dag}$ & $0.916^{\dag}$ & $0.919^{\dag}$ & $0.930^{\dag}$ & $0.939^{\dag}$ & $0.941^{\dag}$\\
& SCHI & $0.817^{\ddag}$ & $0.883^{\dag}$ & $0.913^{\dag}$ & $0.927^{\dag}$ & $0.941^{\dag}$ & $0.948^{\dag}$ & $0.954^{\dag}$ & $0.960^{\dag}$ & $0.963^{\dag}$ & $0.966^{\dag}$\\
& SSU & $\textbf{0.826}^{\ddag}$ & $0.880^{\dag}$ & $0.905^{\dag}$ & $0.922^{\dag}$ & $0.933^{\dag}$ & $0.940^{\dag}$ & $0.949^{\dag}$ & $0.953^{\dag}$ & $0.958^{\dag}$ & $0.961^{\dag}$\\
& CHI & $0.780^{\dag}$ & $0.828^{\dag}$ & $0.891^{\dag}$ & $0.914^{\dag}$ & $0.924^{\dag}$ & $0.933^{\dag}$ & $0.940^{\dag}$ & $0.947^{\dag}$ & $0.952^{\dag}$ & $0.953^{\dag}$\\
& SU & $0.731^{\dag}$ & $0.796^{\dag}$ & $0.832^{\dag}$ & $0.863^{\dag}$ & $0.877^{\dag}$ & $0.910^{\dag}$ & $0.922^{\dag}$ & $0.934^{\dag}$ & $0.942^{\dag}$ & $0.946^{\dag}$\\
& ReliefF & $0.721^{\dag}$ & $0.798^{\dag}$ & $0.841^{\dag}$ & $0.869^{\dag}$ & $0.890^{\dag}$ & $0.904^{\dag}$ & $0.917^{\dag}$ & $0.928^{\dag}$ & $0.935^{\dag}$ & $0.942^{\dag}$\\
\bottomrule
\end{tabular}
\end{lrbox}
\scalebox{0.85}{\usebox{\ISOLET}}
\end{table}
\begin{table}
\centering
\caption{Average MAUC obtained with the nine compared methods on the MNIST data set. The results were obtained by repeating 10-fold cross-validation procedure for 10 times. For each classifier and feature subset size, the Wilcoxon signed-rank test with 95\% confidence level is employed to compare MDFS and 8 other methods. The methods that performed significantly worse (better) than MDFS are highlighted with \dag (\ddag). No superscript is used if no statistical significant difference is detected. The largest MAUC value of each configuration is in boldface.}
\newsavebox{\MNIST}
\begin{lrbox}{\MNIST}
\begin{tabular}{lr*{10}{c}}
\toprule
& & \multicolumn{10}{c}{Feature Subset Size}\\
\cmidrule(r){3-12}
& & 10 & 20 & 30 & 40 & 50 & 60 & 70 & 80 & 90 & 100\\
\midrule
\multirow{9}{*}{Naive Bayes}
& MDFS & \textbf{0.889} & \textbf{0.929} & \textbf{0.944} & \textbf{0.952} & \textbf{0.956} & \textbf{0.961} & \textbf{0.963} & \textbf{0.965} & \textbf{0.967} & \textbf{0.968}\\
& MAUCD & $0.828^{\dag}$ & $0.889^{\dag}$ & $0.922^{\dag}$ & $0.936^{\dag}$ & $0.944^{\dag}$ & $0.950^{\dag}$ & $0.955^{\dag}$ & $0.957^{\dag}$ & $0.958^{\dag}$ & $0.960^{\dag}$\\
& mRMR & $0.877^{\dag}$ & $0.914^{\dag}$ & $0.933^{\dag}$ & $0.939^{\dag}$ & $0.945^{\dag}$ & $0.950^{\dag}$ & $0.952^{\dag}$ & $0.954^{\dag}$ & $0.954^{\dag}$ & $0.956^{\dag}$\\
& FSDD & $0.500^{\dag}$ & $0.500^{\dag}$ & $0.500^{\dag}$ & $0.720^{\dag}$ & $0.732^{\dag}$ & $0.754^{\dag}$ & $0.793^{\dag}$ & $0.806^{\dag}$ & $0.802^{\dag}$ & $0.795^{\dag}$\\
& SCHI & $0.803^{\dag}$ & $0.842^{\dag}$ & $0.871^{\dag}$ & $0.886^{\dag}$ & $0.906^{\dag}$ & $0.914^{\dag}$ & $0.919^{\dag}$ & $0.923^{\dag}$ & $0.927^{\dag}$ & $0.931^{\dag}$\\
& SSU & $0.826^{\dag}$ & $0.870^{\dag}$ & $0.888^{\dag}$ & $0.901^{\dag}$ & $0.915^{\dag}$ & $0.922^{\dag}$ & $0.926^{\dag}$ & $0.929^{\dag}$ & $0.932^{\dag}$ & $0.935^{\dag}$\\
& CHI & $0.834^{\dag}$ & $0.879^{\dag}$ & $0.914^{\dag}$ & $0.928^{\dag}$ & $0.933^{\dag}$ & $0.939^{\dag}$ & $0.941^{\dag}$ & $0.941^{\dag}$ & $0.944^{\dag}$ & $0.947^{\dag}$\\
& SU & $0.844^{\dag}$ & $0.891^{\dag}$ & $0.918^{\dag}$ & $0.932^{\dag}$ & $0.939^{\dag}$ & $0.941^{\dag}$ & $0.942^{\dag}$ & $0.946^{\dag}$ & $0.949^{\dag}$ & $0.950^{\dag}$\\
& ReliefF & $0.843^{\dag}$ & $0.892^{\dag}$ & $0.917^{\dag}$ & $0.933^{\dag}$ & $0.943^{\dag}$ & $0.950^{\dag}$ & $0.955^{\dag}$ & $0.959^{\dag}$ & $0.962^{\dag}$ & $0.964^{\dag}$\\
\midrule
\multirow{9}{*}{C4.5}
& MDFS & \textbf{0.829} & \textbf{0.865} & \textbf{0.878} & \textbf{0.884} & \textbf{0.888} & \textbf{0.893} & \textbf{0.894} & \textbf{0.898} & \textbf{0.900} & \textbf{0.901}\\
& MAUCD & $0.766^{\dag}$ & $0.836^{\dag}$ & $0.863^{\dag}$ & $0.872^{\dag}$ & $0.877^{\dag}$ & $0.880^{\dag}$ & $0.886^{\dag}$ & $0.889^{\dag}$ & $0.889^{\dag}$ & $0.891^{\dag}$\\
& mRMR & $0.809^{\dag}$ & $0.851^{\dag}$ & $0.866^{\dag}$ & $0.876^{\dag}$ & $0.885^{\dag}$ & $0.890^{\dag}$ & $0.893$ & $0.895^{\dag}$ & $0.897^{\dag}$ & $0.896^{\dag}$\\
& FSDD & $0.500^{\dag}$ & $0.500^{\dag}$ & $0.500^{\dag}$ & $0.697^{\dag}$ & $0.725^{\dag}$ & $0.740^{\dag}$ & $0.764^{\dag}$ & $0.769^{\dag}$ & $0.768^{\dag}$ & $0.768^{\dag}$\\
& SCHI & $0.818^{\dag}$ & $0.839^{\dag}$ & $0.852^{\dag}$ & $0.860^{\dag}$ & $0.871^{\dag}$ & $0.878^{\dag}$ & $0.880^{\dag}$ & $0.882^{\dag}$ & $0.888^{\dag}$ & $0.890^{\dag}$\\
& SSU & $0.807^{\dag}$ & $0.838^{\dag}$ & $0.853^{\dag}$ & $0.872^{\dag}$ & $0.883^{\dag}$ & $0.889^{\dag}$ & $0.890^{\dag}$ & $0.893^{\dag}$ & $0.894^{\dag}$ & $0.897^{\dag}$\\
& CHI & $0.784^{\dag}$ & $0.824^{\dag}$ & $0.852^{\dag}$ & $0.869^{\dag}$ & $0.878^{\dag}$ & $0.883^{\dag}$ & $0.884^{\dag}$ & $0.884^{\dag}$ & $0.888^{\dag}$ & $0.892^{\dag}$\\
& SU & $0.804^{\dag}$ & $0.838^{\dag}$ & $0.864^{\dag}$ & $0.872^{\dag}$ & $0.879^{\dag}$ & $0.883^{\dag}$ & $0.885^{\dag}$ & $0.888^{\dag}$ & $0.890^{\dag}$ & $0.892^{\dag}$\\
& ReliefF & $0.796^{\dag}$ & $0.837^{\dag}$ & $0.859^{\dag}$ & $0.871^{\dag}$ & $0.881^{\dag}$ & $0.885^{\dag}$ & $0.889^{\dag}$ & $0.892^{\dag}$ & $0.895^{\dag}$ & $0.896^{\dag}$\\
\midrule
\multirow{9}{*}{1NN}
& MDFS & \textbf{0.764} & \textbf{0.855} & \textbf{0.892} & \textbf{0.912} & \textbf{0.927} & \textbf{0.938} & \textbf{0.947} & \textbf{0.952} & \textbf{0.956} & \textbf{0.959}\\
& MAUCD & $0.712^{\dag}$ & $0.809^{\dag}$ & $0.860^{\dag}$ & $0.886^{\dag}$ & $0.897^{\dag}$ & $0.910^{\dag}$ & $0.922^{\dag}$ & $0.925^{\dag}$ & $0.930^{\dag}$ & $0.935^{\dag}$\\
& mRMR & $0.761^{\dag}$ & $0.842^{\dag}$ & $0.879^{\dag}$ & $0.904^{\dag}$ & $0.920^{\dag}$ & $0.930^{\dag}$ & $0.936^{\dag}$ & $0.941^{\dag}$ & $0.945^{\dag}$ & $0.949^{\dag}$\\
& FSDD & $0.500^{\dag}$ & $0.500^{\dag}$ & $0.500^{\dag}$ & $0.595^{\dag}$ & $0.646^{\dag}$ & $0.678^{\dag}$ & $0.714^{\dag}$ & $0.726^{\dag}$ & $0.728^{\dag}$ & $0.728^{\dag}$\\
& SCHI & $0.679^{\dag}$ & $0.762^{\dag}$ & $0.814^{\dag}$ & $0.843^{\dag}$ & $0.872^{\dag}$ & $0.889^{\dag}$ & $0.901^{\dag}$ & $0.911^{\dag}$ & $0.919^{\dag}$ & $0.928^{\dag}$\\
& SSU & $0.674^{\dag}$ & $0.749^{\dag}$ & $0.792^{\dag}$ & $0.834^{\dag}$ & $0.872^{\dag}$ & $0.895^{\dag}$ & $0.906^{\dag}$ & $0.914^{\dag}$ & $0.924^{\dag}$ & $0.931^{\dag}$\\
& CHI & $0.725^{\dag}$ & $0.800^{\dag}$ & $0.853^{\dag}$ & $0.879^{\dag}$ & $0.898^{\dag}$ & $0.913^{\dag}$ & $0.919^{\dag}$ & $0.923^{\dag}$ & $0.927^{\dag}$ & $0.934^{\dag}$\\
& SU & $0.737^{\dag}$ & $0.810^{\dag}$ & $0.852^{\dag}$ & $0.880^{\dag}$ & $0.899^{\dag}$ & $0.911^{\dag}$ & $0.916^{\dag}$ & $0.923^{\dag}$ & $0.927^{\dag}$ & $0.933^{\dag}$\\
& ReliefF & $0.727^{\dag}$ & $0.823^{\dag}$ & $0.873^{\dag}$ & $0.902^{\dag}$ & $0.919^{\dag}$ & $0.931^{\dag}$ & $0.939^{\dag}$ & $0.947^{\dag}$ & $0.952^{\dag}$ & $0.956^{\dag}$\\
\midrule
\multirow{9}{*}{SVM}
& MDFS & \textbf{0.786} & \textbf{0.878} & \textbf{0.914} & \textbf{0.930} & \textbf{0.939} & \textbf{0.945} & \textbf{0.948} & \textbf{0.951} & 0.952 & 0.953\\
& MAUCD & $0.738^{\dag}$ & $0.836^{\dag}$ & $0.883^{\dag}$ & $0.908^{\dag}$ & $0.918^{\dag}$ & $0.925^{\dag}$ & $0.935^{\dag}$ & $0.938^{\dag}$ & $0.941^{\dag}$ & $0.944^{\dag}$\\
& mRMR & $0.778^{\dag}$ & $0.868^{\dag}$ & $0.904^{\dag}$ & $0.924^{\dag}$ & $0.936^{\dag}$ & $0.944$ & $\textbf{0.948}$ & $\textbf{0.951}$ & $\textbf{0.953}$ & $\textbf{0.955}^{\ddag}$\\
& FSDD & $0.500^{\dag}$ & $0.500^{\dag}$ & $0.500^{\dag}$ & $0.648^{\dag}$ & $0.693^{\dag}$ & $0.715^{\dag}$ & $0.746^{\dag}$ & $0.757^{\dag}$ & $0.759^{\dag}$ & $0.760^{\dag}$\\
& SCHI & $0.737^{\dag}$ & $0.796^{\dag}$ & $0.843^{\dag}$ & $0.873^{\dag}$ & $0.907^{\dag}$ & $0.923^{\dag}$ & $0.933^{\dag}$ & $0.941^{\dag}$ & $0.948^{\dag}$ & $0.953$\\
& SSU & $0.725^{\dag}$ & $0.788^{\dag}$ & $0.824^{\dag}$ & $0.866^{\dag}$ & $0.904^{\dag}$ & $0.927^{\dag}$ & $0.937^{\dag}$ & $0.944^{\dag}$ & $0.950^{\dag}$ & $\textbf{0.955}^{\ddag}$\\
& CHI & $0.755^{\dag}$ & $0.826^{\dag}$ & $0.882^{\dag}$ & $0.906^{\dag}$ & $0.919^{\dag}$ & $0.931^{\dag}$ & $0.934^{\dag}$ & $0.937^{\dag}$ & $0.940^{\dag}$ & $0.944^{\dag}$\\
& SU & $0.763^{\dag}$ & $0.836^{\dag}$ & $0.883^{\dag}$ & $0.909^{\dag}$ & $0.923^{\dag}$ & $0.932^{\dag}$ & $0.935^{\dag}$ & $0.940^{\dag}$ & $0.942^{\dag}$ & $0.946^{\dag}$\\
& ReliefF & $0.752^{\dag}$ & $0.850^{\dag}$ & $0.900^{\dag}$ & $0.922^{\dag}$ & $0.933^{\dag}$ & $0.940^{\dag}$ & $0.944^{\dag}$ & $0.947^{\dag}$ & $0.949^{\dag}$ & $0.950^{\dag}$\\
\bottomrule
\end{tabular}
\end{lrbox}
\scalebox{0.85}{\usebox{\MNIST}}
\end{table}
\begin{table}
\centering
\caption{Average MAUC obtained with the nine compared methods on the USPS data set. The results were obtained by repeating 10-fold cross-validation procedure for 10 times. For each classifier and feature subset size, the Wilcoxon signed-rank test with 95\% confidence level is employed to compare MDFS and 8 other methods. The methods that performed significantly worse (better) than MDFS are highlighted with \dag (\ddag). No superscript is used if no statistical significant difference is detected. The largest MAUC value of each configuration is in boldface.}
\newsavebox{\USPS}
\begin{lrbox}{\USPS}
\begin{tabular}{lr*{10}{c}}
\toprule
& & \multicolumn{10}{c}{Feature Subset Size}\\
\cmidrule(r){3-12}
& & 10 & 20 & 30 & 40 & 50 & 60 & 70 & 80 & 90 & 100\\
\midrule
\multirow{9}{*}{Naive Bayes}
& MDFS & \textbf{0.929} & \textbf{0.962} & \textbf{0.972} & \textbf{0.976} & \textbf{0.978} & \textbf{0.979} & \textbf{0.980} & \textbf{0.981} & \textbf{0.981} & \textbf{0.981}\\
& MAUCD & $0.912^{\dag}$ & $0.928^{\dag}$ & $0.948^{\dag}$ & $0.964^{\dag}$ & $0.966^{\dag}$ & $0.971^{\dag}$ & $0.971^{\dag}$ & $0.974^{\dag}$ & $0.975^{\dag}$ & $0.977^{\dag}$\\
& mRMR & $0.921^{\dag}$ & $0.947^{\dag}$ & $0.960^{\dag}$ & $0.965^{\dag}$ & $0.967^{\dag}$ & $0.973^{\dag}$ & $0.975^{\dag}$ & $0.976^{\dag}$ & $0.977^{\dag}$ & $0.978^{\dag}$\\
& FSDD & $0.914^{\dag}$ & $0.929^{\dag}$ & $0.944^{\dag}$ & $0.951^{\dag}$ & $0.967^{\dag}$ & $0.968^{\dag}$ & $0.969^{\dag}$ & $0.972^{\dag}$ & $0.974^{\dag}$ & $0.975^{\dag}$\\
& SCHI & $0.907^{\dag}$ & $0.953^{\dag}$ & $0.968^{\dag}$ & $0.970^{\dag}$ & $0.971^{\dag}$ & $0.972^{\dag}$ & $0.972^{\dag}$ & $0.972^{\dag}$ & $0.972^{\dag}$ & $0.971^{\dag}$\\
& SSU & $0.924^{\dag}$ & $0.945^{\dag}$ & $0.955^{\dag}$ & $0.961^{\dag}$ & $0.962^{\dag}$ & $0.963^{\dag}$ & $0.965^{\dag}$ & $0.965^{\dag}$ & $0.965^{\dag}$ & $0.965^{\dag}$\\
& CHI & $0.896^{\dag}$ & $0.923^{\dag}$ & $0.950^{\dag}$ & $0.963^{\dag}$ & $0.966^{\dag}$ & $0.967^{\dag}$ & $0.969^{\dag}$ & $0.973^{\dag}$ & $0.974^{\dag}$ & $0.976^{\dag}$\\
& SU & $0.817^{\dag}$ & $0.926^{\dag}$ & $0.937^{\dag}$ & $0.943^{\dag}$ & $0.951^{\dag}$ & $0.956^{\dag}$ & $0.963^{\dag}$ & $0.967^{\dag}$ & $0.969^{\dag}$ & $0.970^{\dag}$\\
& ReliefF & $0.898^{\dag}$ & $0.938^{\dag}$ & $0.954^{\dag}$ & $0.963^{\dag}$ & $0.968^{\dag}$ & $0.971^{\dag}$ & $0.973^{\dag}$ & $0.975^{\dag}$ & $0.977^{\dag}$ & $0.978^{\dag}$\\
\midrule
\multirow{9}{*}{C4.5}
& MDFS & 0.879 & \textbf{0.908} & \textbf{0.920} & \textbf{0.923} & \textbf{0.927} & \textbf{0.928} & \textbf{0.930} & \textbf{0.932} & \textbf{0.933} & \textbf{0.933}\\
& MAUCD & $0.862^{\dag}$ & $0.873^{\dag}$ & $0.891^{\dag}$ & $0.915^{\dag}$ & $0.917^{\dag}$ & $0.922^{\dag}$ & $0.929$ & $\textbf{0.932}$ & $0.931^{\dag}$ & $0.932$\\
& mRMR & $0.859^{\dag}$ & $0.893^{\dag}$ & $0.895^{\dag}$ & $0.904^{\dag}$ & $0.911^{\dag}$ & $0.921^{\dag}$ & $0.928^{\dag}$ & $0.930^{\dag}$ & $0.931^{\dag}$ & $\textbf{0.933}$\\
& FSDD & $0.862^{\dag}$ & $0.874^{\dag}$ & $0.887^{\dag}$ & $0.903^{\dag}$ & $0.923^{\dag}$ & $0.926$ & $0.928^{\dag}$ & $\textbf{0.932}$ & $0.931$ & $\textbf{0.933}$\\
& SCHI & $0.860^{\dag}$ & $0.901^{\dag}$ & $0.911^{\dag}$ & $0.918^{\dag}$ & $0.925$ & $0.925^{\dag}$ & $0.925^{\dag}$ & $0.925^{\dag}$ & $0.925^{\dag}$ & $0.928^{\dag}$\\
& SSU & $\textbf{0.886}^{\ddag}$ & $0.895^{\dag}$ & $0.904^{\dag}$ & $0.910^{\dag}$ & $0.915^{\dag}$ & $0.918^{\dag}$ & $0.923^{\dag}$ & $0.926^{\dag}$ & $0.927^{\dag}$ & $0.930^{\dag}$\\
& CHI & $0.853^{\dag}$ & $0.873^{\dag}$ & $0.898^{\dag}$ & $0.918^{\dag}$ & $0.924^{\dag}$ & $0.924^{\dag}$ & $0.928^{\dag}$ & $0.930^{\dag}$ & $0.930^{\dag}$ & $0.931^{\dag}$\\
& SU & $0.763^{\dag}$ & $0.892^{\dag}$ & $0.899^{\dag}$ & $0.902^{\dag}$ & $0.909^{\dag}$ & $0.914^{\dag}$ & $0.922^{\dag}$ & $0.926^{\dag}$ & $0.927^{\dag}$ & $0.928^{\dag}$\\
& ReliefF & $0.853^{\dag}$ & $0.883^{\dag}$ & $0.896^{\dag}$ & $0.907^{\dag}$ & $0.912^{\dag}$ & $0.917^{\dag}$ & $0.921^{\dag}$ & $0.925^{\dag}$ & $0.926^{\dag}$ & $0.928^{\dag}$\\
\midrule
\multirow{9}{*}{1NN}
& MDFS & \textbf{0.813} & \textbf{0.905} & \textbf{0.939} & \textbf{0.954} & \textbf{0.963} & \textbf{0.968} & \textbf{0.972} & \textbf{0.975} & \textbf{0.976} & \textbf{0.977}\\
& MAUCD & $0.790^{\dag}$ & $0.844^{\dag}$ & $0.885^{\dag}$ & $0.926^{\dag}$ & $0.937^{\dag}$ & $0.950^{\dag}$ & $0.957^{\dag}$ & $0.965^{\dag}$ & $0.969^{\dag}$ & $0.973^{\dag}$\\
& mRMR & $0.796^{\dag}$ & $0.884^{\dag}$ & $0.915^{\dag}$ & $0.935^{\dag}$ & $0.945^{\dag}$ & $0.957^{\dag}$ & $0.965^{\dag}$ & $0.970^{\dag}$ & $0.972^{\dag}$ & $0.975^{\dag}$\\
& FSDD & $0.787^{\dag}$ & $0.846^{\dag}$ & $0.875^{\dag}$ & $0.913^{\dag}$ & $0.947^{\dag}$ & $0.955^{\dag}$ & $0.961^{\dag}$ & $0.968^{\dag}$ & $0.972^{\dag}$ & $0.974^{\dag}$\\
& SCHI & $0.782^{\dag}$ & $0.890^{\dag}$ & $0.931^{\dag}$ & $0.947^{\dag}$ & $0.954^{\dag}$ & $0.959^{\dag}$ & $0.961^{\dag}$ & $0.965^{\dag}$ & $0.967^{\dag}$ & $0.968^{\dag}$\\
& SSU & $0.807^{\dag}$ & $0.872^{\dag}$ & $0.912^{\dag}$ & $0.935^{\dag}$ & $0.945^{\dag}$ & $0.953^{\dag}$ & $0.961^{\dag}$ & $0.966^{\dag}$ & $0.969^{\dag}$ & $0.972^{\dag}$\\
& CHI & $0.761^{\dag}$ & $0.832^{\dag}$ & $0.888^{\dag}$ & $0.932^{\dag}$ & $0.944^{\dag}$ & $0.951^{\dag}$ & $0.959^{\dag}$ & $0.966^{\dag}$ & $0.969^{\dag}$ & $0.972^{\dag}$\\
& SU & $0.656^{\dag}$ & $0.841^{\dag}$ & $0.876^{\dag}$ & $0.895^{\dag}$ & $0.916^{\dag}$ & $0.928^{\dag}$ & $0.949^{\dag}$ & $0.958^{\dag}$ & $0.963^{\dag}$ & $0.967^{\dag}$\\
& ReliefF & $0.786^{\dag}$ & $0.874^{\dag}$ & $0.913^{\dag}$ & $0.933^{\dag}$ & $0.946^{\dag}$ & $0.955^{\dag}$ & $0.961^{\dag}$ & $0.966^{\dag}$ & $0.970^{\dag}$ & $0.972^{\dag}$\\
\midrule
\multirow{9}{*}{SVM}
& MDFS & 0.829 & \textbf{0.924} & \textbf{0.957} & \textbf{0.968} & \textbf{0.975} & \textbf{0.978} & \textbf{0.980} & \textbf{0.982} & \textbf{0.984} & \textbf{0.985}\\
& MAUCD & $0.817^{\dag}$ & $0.868^{\dag}$ & $0.913^{\dag}$ & $0.948^{\dag}$ & $0.958^{\dag}$ & $0.967^{\dag}$ & $0.972^{\dag}$ & $0.979^{\dag}$ & $0.982^{\dag}$ & $0.984^{\dag}$\\
& mRMR & $0.820^{\dag}$ & $0.904^{\dag}$ & $0.939^{\dag}$ & $0.958^{\dag}$ & $0.966^{\dag}$ & $0.973^{\dag}$ & $0.979^{\dag}$ & $0.981^{\dag}$ & $0.982^{\dag}$ & $0.984^{\dag}$\\
& FSDD & $0.817^{\dag}$ & $0.866^{\dag}$ & $0.904^{\dag}$ & $0.938^{\dag}$ & $0.965^{\dag}$ & $0.971^{\dag}$ & $0.975^{\dag}$ & $0.980^{\dag}$ & $0.982^{\dag}$ & $0.983^{\dag}$\\
& SCHI & $0.809^{\dag}$ & $0.916^{\dag}$ & $0.952^{\dag}$ & $0.965^{\dag}$ & $0.969^{\dag}$ & $0.973^{\dag}$ & $0.977^{\dag}$ & $0.978^{\dag}$ & $0.980^{\dag}$ & $0.981^{\dag}$\\
& SSU & $\textbf{0.835}^{\ddag}$ & $0.902^{\dag}$ & $0.936^{\dag}$ & $0.955^{\dag}$ & $0.964^{\dag}$ & $0.971^{\dag}$ & $0.974^{\dag}$ & $0.978^{\dag}$ & $0.980^{\dag}$ & $0.982^{\dag}$\\
& CHI & $0.793^{\dag}$ & $0.860^{\dag}$ & $0.912^{\dag}$ & $0.953^{\dag}$ & $0.963^{\dag}$ & $0.969^{\dag}$ & $0.974^{\dag}$ & $0.979^{\dag}$ & $0.981^{\dag}$ & $0.983^{\dag}$\\
& SU & $0.681^{\dag}$ & $0.877^{\dag}$ & $0.907^{\dag}$ & $0.923^{\dag}$ & $0.939^{\dag}$ & $0.950^{\dag}$ & $0.968^{\dag}$ & $0.974^{\dag}$ & $0.976^{\dag}$ & $0.979^{\dag}$\\
& ReliefF & $0.810^{\dag}$ & $0.893^{\dag}$ & $0.933^{\dag}$ & $0.951^{\dag}$ & $0.963^{\dag}$ & $0.970^{\dag}$ & $0.975^{\dag}$ & $0.979^{\dag}$ & $0.981^{\dag}$ & $0.983^{\dag}$\\
\bottomrule
\end{tabular}
\end{lrbox}
\scalebox{0.85}{\usebox{\USPS}}
\end{table}
\begin{table}
\centering
\caption{Average MAUC obtained with the nine compared methods on the Phoneme data set. The results were obtained by repeating 10-fold cross-validation procedure for 10 times. For each classifier and feature subset size, the Wilcoxon signed-rank test with 95\% confidence level is employed to compare MDFS and 8 other methods. The methods that performed significantly worse (better) than MDFS are highlighted with \dag (\ddag). No superscript is used if no statistical significant difference is detected. The largest MAUC value of each configuration is in boldface.}
\newsavebox{\Phoneme}
\begin{lrbox}{\Phoneme}
\begin{tabular}{lr*{10}{c}}
\toprule
& & \multicolumn{10}{c}{Feature Subset Size}\\
\cmidrule(r){3-12}
& & 10 & 20 & 30 & 40 & 50 & 60 & 70 & 80 & 90 & 100\\
\midrule
\multirow{9}{*}{Naive Bayes}
& MDFS & 0.974 & 0.980 & 0.982 & \textbf{0.983} & 0.983 & 0.983 & 0.983 & \textbf{0.983} & \textbf{0.983} & \textbf{0.983}\\
& MAUCD & $0.956^{\dag}$ & $0.959^{\dag}$ & $0.977^{\dag}$ & $0.981^{\dag}$ & $0.980^{\dag}$ & $0.980^{\dag}$ & $0.979^{\dag}$ & $0.978^{\dag}$ & $0.977^{\dag}$ & $0.976^{\dag}$\\
& mRMR & $0.977^{\ddag}$ & $0.979^{\dag}$ & $0.979^{\dag}$ & $0.980^{\dag}$ & $0.981^{\dag}$ & $0.981^{\dag}$ & $0.981^{\dag}$ & $0.981^{\dag}$ & $0.981^{\dag}$ & $0.981^{\dag}$\\
& FSDD & $0.937^{\dag}$ & $0.954^{\dag}$ & $0.977^{\dag}$ & $0.982^{\dag}$ & $0.983$ & $0.982^{\dag}$ & $0.981^{\dag}$ & $0.981^{\dag}$ & $0.980^{\dag}$ & $0.979^{\dag}$\\
& SCHI & $\textbf{0.980}^{\ddag}$ & $\textbf{0.982}^{\ddag}$ & $\textbf{0.983}^{\ddag}$ & $\textbf{0.983}$ & $0.983$ & $0.983$ & $0.983^{\dag}$ & $\textbf{0.983}^{\dag}$ & $0.982^{\dag}$ & $0.982^{\dag}$\\
& SSU & $0.976$ & $0.980$ & $0.982$ & $0.982^{\dag}$ & $0.982^{\dag}$ & $0.982^{\dag}$ & $0.982^{\dag}$ & $0.982^{\dag}$ & $0.982^{\dag}$ & $0.982^{\dag}$\\
& CHI & $0.946^{\dag}$ & $0.955^{\dag}$ & $0.970^{\dag}$ & $0.982^{\dag}$ & $0.981^{\dag}$ & $0.980^{\dag}$ & $0.979^{\dag}$ & $0.977^{\dag}$ & $0.976^{\dag}$ & $0.974^{\dag}$\\
& SU & $0.936^{\dag}$ & $0.954^{\dag}$ & $0.978^{\dag}$ & $0.982$ & $0.982^{\dag}$ & $0.981^{\dag}$ & $0.979^{\dag}$ & $0.978^{\dag}$ & $0.976^{\dag}$ & $0.975^{\dag}$\\
& ReliefF & $0.954^{\dag}$ & $0.965^{\dag}$ & $0.975^{\dag}$ & $0.982^{\dag}$ & $\textbf{0.984}^{\ddag}$ & $\textbf{0.984}^{\ddag}$ & $\textbf{0.984}^{\ddag}$ & $\textbf{0.983}$ & $\textbf{0.983}$ & $\textbf{0.983}^{\dag}$\\
\midrule
\multirow{9}{*}{C4.5}
& MDFS & 0.945 & \textbf{0.947} & \textbf{0.944} & 0.939 & 0.938 & 0.936 & 0.933 & 0.932 & 0.931 & 0.930\\
& MAUCD & $0.925^{\dag}$ & $0.921^{\dag}$ & $0.938^{\dag}$ & $0.937$ & $0.937$ & $0.932^{\dag}$ & $0.931^{\dag}$ & $0.931$ & $0.928$ & $0.928$\\
& mRMR & $\textbf{0.950}^{\ddag}$ & $0.946$ & $0.939^{\dag}$ & $0.937$ & $0.933^{\dag}$ & $0.930^{\dag}$ & $0.932$ & $0.933$ & $0.932$ & $0.930$\\
& FSDD & $0.902^{\dag}$ & $0.909^{\dag}$ & $0.938^{\dag}$ & $0.938$ & $\textbf{0.939}$ & $\textbf{0.938}$ & $0.936^{\ddag}$ & $0.933$ & $0.933$ & $0.931$\\
& SCHI & $0.949^{\ddag}$ & $0.943^{\dag}$ & $0.939^{\dag}$ & $0.935^{\dag}$ & $0.933^{\dag}$ & $0.933$ & $0.933$ & $0.931$ & $0.931$ & $0.929$\\
& SSU & $0.945$ & $0.944^{\dag}$ & $0.937^{\dag}$ & $0.936^{\dag}$ & $0.931^{\dag}$ & $0.930^{\dag}$ & $0.932$ & $0.929^{\dag}$ & $0.929$ & $0.928$\\
& CHI & $0.913^{\dag}$ & $0.914^{\dag}$ & $0.925^{\dag}$ & $0.935^{\dag}$ & $0.932^{\dag}$ & $0.929^{\dag}$ & $0.928^{\dag}$ & $0.927^{\dag}$ & $0.925^{\dag}$ & $0.923^{\dag}$\\
& SU & $0.901^{\dag}$ & $0.910^{\dag}$ & $0.939^{\dag}$ & $0.937^{\dag}$ & $0.936$ & $0.933$ & $0.930^{\dag}$ & $0.927^{\dag}$ & $0.927^{\dag}$ & $0.929$\\
& ReliefF & $0.923^{\dag}$ & $0.925^{\dag}$ & $0.935^{\dag}$ & $\textbf{0.941}$ & $0.938$ & $0.937$ & $\textbf{0.937}^{\ddag}$ & $\textbf{0.934}$ & $\textbf{0.934}$ & $\textbf{0.932}$\\
\midrule
\multirow{9}{*}{1NN}
& MDFS & 0.903 & \textbf{0.923} & \textbf{0.926} & 0.927 & 0.928 & 0.929 & 0.929 & 0.928 & 0.928 & 0.928\\
& MAUCD & $0.865^{\dag}$ & $0.892^{\dag}$ & $0.917^{\dag}$ & $0.926$ & $0.922^{\dag}$ & $0.921^{\dag}$ & $0.920^{\dag}$ & $0.920^{\dag}$ & $0.920^{\dag}$ & $0.918^{\dag}$\\
& mRMR & $0.905$ & $0.919^{\dag}$ & $0.921^{\dag}$ & $0.927$ & $0.926^{\dag}$ & $0.927$ & $0.929$ & $\textbf{0.931}^{\ddag}$ & $\textbf{0.933}^{\ddag}$ & $\textbf{0.932}^{\ddag}$\\
& FSDD & $0.849^{\dag}$ & $0.882^{\dag}$ & $0.917^{\dag}$ & $\textbf{0.931}^{\ddag}$ & $\textbf{0.933}^{\ddag}$ & $\textbf{0.932}^{\ddag}$ & $0.929$ & $0.925^{\dag}$ & $0.925^{\dag}$ & $0.926$\\
& SCHI & $\textbf{0.906}$ & $0.920^{\dag}$ & $0.922^{\dag}$ & $0.923^{\dag}$ & $0.926$ & $0.928$ & $0.929$ & $\textbf{0.931}^{\ddag}$ & $0.930$ & $0.931^{\ddag}$\\
& SSU & $0.898^{\dag}$ & $0.918^{\dag}$ & $0.921^{\dag}$ & $0.925$ & $0.925^{\dag}$ & $0.926^{\dag}$ & $0.928$ & $0.930$ & $0.931^{\ddag}$ & $0.931^{\ddag}$\\
& CHI & $0.858^{\dag}$ & $0.884^{\dag}$ & $0.890^{\dag}$ & $0.925$ & $0.923^{\dag}$ & $0.923^{\dag}$ & $0.920^{\dag}$ & $0.917^{\dag}$ & $0.915^{\dag}$ & $0.913^{\dag}$\\
& SU & $0.843^{\dag}$ & $0.882^{\dag}$ & $0.919^{\dag}$ & $0.930^{\ddag}$ & $0.928$ & $0.926^{\dag}$ & $0.923^{\dag}$ & $0.921^{\dag}$ & $0.920^{\dag}$ & $0.919^{\dag}$\\
& ReliefF & $0.872^{\dag}$ & $0.901^{\dag}$ & $0.919^{\dag}$ & $0.929^{\ddag}$ & $0.931^{\ddag}$ & $\textbf{0.932}^{\ddag}$ & $\textbf{0.932}^{\ddag}$ & $\textbf{0.931}^{\ddag}$ & $0.931^{\ddag}$ & $0.930^{\ddag}$\\
\midrule
\multirow{9}{*}{SVM}
& MDFS & 0.917 & \textbf{0.939} & \textbf{0.943} & 0.946 & 0.948 & 0.949 & 0.950 & 0.950 & 0.950 & \textbf{0.951}\\
& MAUCD & $0.867^{\dag}$ & $0.904^{\dag}$ & $0.933^{\dag}$ & $0.938^{\dag}$ & $0.940^{\dag}$ & $0.943^{\dag}$ & $0.943^{\dag}$ & $0.944^{\dag}$ & $0.944^{\dag}$ & $0.945^{\dag}$\\
& mRMR & $0.925^{\ddag}$ & $0.935^{\dag}$ & $0.937^{\dag}$ & $0.939^{\dag}$ & $0.941^{\dag}$ & $0.943^{\dag}$ & $0.945^{\dag}$ & $0.947^{\dag}$ & $0.949$ & $\textbf{0.951}$\\
& FSDD & $0.820^{\dag}$ & $0.890^{\dag}$ & $0.933^{\dag}$ & $\textbf{0.947}$ & $\textbf{0.949}$ & $0.949$ & $0.949$ & $0.949$ & $0.949$ & $0.949$\\
& SCHI & $\textbf{0.926}^{\ddag}$ & $0.935^{\dag}$ & $0.939^{\dag}$ & $0.942^{\dag}$ & $0.945^{\dag}$ & $0.947^{\dag}$ & $0.948^{\dag}$ & $0.949^{\dag}$ & $0.950$ & $0.950$\\
& SSU & $0.919$ & $0.934^{\dag}$ & $0.938^{\dag}$ & $0.942^{\dag}$ & $0.944^{\dag}$ & $0.945^{\dag}$ & $0.948^{\dag}$ & $0.949^{\dag}$ & $0.950$ & $\textbf{0.951}$\\
& CHI & $0.860^{\dag}$ & $0.894^{\dag}$ & $0.918^{\dag}$ & $0.939^{\dag}$ & $0.940^{\dag}$ & $0.942^{\dag}$ & $0.942^{\dag}$ & $0.942^{\dag}$ & $0.943^{\dag}$ & $0.943^{\dag}$\\
& SU & $0.830^{\dag}$ & $0.892^{\dag}$ & $0.935^{\dag}$ & $0.943^{\dag}$ & $0.945^{\dag}$ & $0.945^{\dag}$ & $0.945^{\dag}$ & $0.946^{\dag}$ & $0.946^{\dag}$ & $0.948^{\dag}$\\
& ReliefF & $0.868^{\dag}$ & $0.911^{\dag}$ & $0.938^{\dag}$ & $0.946$ & $\textbf{0.949}$ & $\textbf{0.950}^{\ddag}$ & $\textbf{0.951}^{\ddag}$ & $\textbf{0.951}$ & $\textbf{0.951}$ & $\textbf{0.951}$\\
\bottomrule
\end{tabular}
\end{lrbox}
\scalebox{0.85}{\usebox{\Phoneme}}
\end{table}
\begin{table}
\centering
\caption{Average MAUC obtained with the nine compared methods on the Washington data set. The results were obtained by repeating 10-fold cross-validation procedure for 10 times. For each classifier and feature subset size, the Wilcoxon signed-rank test with 95\% confidence level is employed to compare MDFS and 8 other methods. The methods that performed significantly worse (better) than MDFS are highlighted with \dag (\ddag). No superscript is used if no statistical significant difference is detected. The largest MAUC value of each configuration is in boldface.}
\newsavebox{\Washington}
\begin{lrbox}{\Washington}
\begin{tabular}{lr*{10}{c}}
\toprule
& & \multicolumn{10}{c}{Feature Subset Size}\\
\cmidrule(r){3-12}
& & 10 & 20 & 30 & 40 & 50 & 60 & 70 & 80 & 90 & 100\\
\midrule
\multirow{9}{*}{Naive Bayes}
& MDFS & \textbf{0.970} & \textbf{0.973} & \textbf{0.973} & \textbf{0.972} & \textbf{0.972} & \textbf{0.971} & \textbf{0.971} & \textbf{0.970} & \textbf{0.970} & \textbf{0.970}\\
& MAUCD & $0.934^{\dag}$ & $0.930^{\dag}$ & $0.927^{\dag}$ & $0.926^{\dag}$ & $0.926^{\dag}$ & $0.938^{\dag}$ & $0.941^{\dag}$ & $0.940^{\dag}$ & $0.943^{\dag}$ & $0.949^{\dag}$\\
& mRMR & $0.968^{\dag}$ & $0.970^{\dag}$ & $0.971^{\dag}$ & $0.970^{\dag}$ & $0.970^{\dag}$ & $0.969^{\dag}$ & $0.969^{\dag}$ & $0.969^{\dag}$ & $0.969^{\dag}$ & $0.969^{\dag}$\\
& FSDD & $0.918^{\dag}$ & $0.934^{\dag}$ & $0.939^{\dag}$ & $0.940^{\dag}$ & $0.939^{\dag}$ & $0.945^{\dag}$ & $0.948^{\dag}$ & $0.947^{\dag}$ & $0.946^{\dag}$ & $0.947^{\dag}$\\
& SCHI & $0.968^{\dag}$ & $0.971^{\dag}$ & $0.971^{\dag}$ & $0.970^{\dag}$ & $0.969^{\dag}$ & $0.968^{\dag}$ & $0.968^{\dag}$ & $0.968^{\dag}$ & $0.968^{\dag}$ & $0.968^{\dag}$\\
& SSU & $0.969^{\dag}$ & $0.971^{\dag}$ & $0.971^{\dag}$ & $0.971^{\dag}$ & $0.971^{\dag}$ & $\textbf{0.971}^{\dag}$ & $0.970^{\dag}$ & $\textbf{0.970}^{\dag}$ & $\textbf{0.970}^{\dag}$ & $0.969^{\dag}$\\
& CHI & $0.946^{\dag}$ & $0.945^{\dag}$ & $0.944^{\dag}$ & $0.942^{\dag}$ & $0.942^{\dag}$ & $0.940^{\dag}$ & $0.950^{\dag}$ & $0.953^{\dag}$ & $0.955^{\dag}$ & $0.956^{\dag}$\\
& SU & $0.944^{\dag}$ & $0.947^{\dag}$ & $0.948^{\dag}$ & $0.949^{\dag}$ & $0.951^{\dag}$ & $0.953^{\dag}$ & $0.955^{\dag}$ & $0.956^{\dag}$ & $0.957^{\dag}$ & $0.957^{\dag}$\\
& ReliefF & $0.899^{\dag}$ & $0.910^{\dag}$ & $0.920^{\dag}$ & $0.927^{\dag}$ & $0.936^{\dag}$ & $0.941^{\dag}$ & $0.947^{\dag}$ & $0.956^{\dag}$ & $0.960^{\dag}$ & $0.962^{\dag}$\\
\midrule
\multirow{9}{*}{C4.5}
& MDFS & 0.961 & 0.955 & 0.952 & 0.949 & 0.947 & 0.947 & 0.946 & \textbf{0.944} & \textbf{0.943} & 0.942\\
& MAUCD & $0.948^{\dag}$ & $0.948^{\dag}$ & $0.941^{\dag}$ & $0.935^{\dag}$ & $0.930^{\dag}$ & $0.933^{\dag}$ & $0.934^{\dag}$ & $0.932^{\dag}$ & $0.932^{\dag}$ & $0.930^{\dag}$\\
& mRMR & $0.960$ & $0.952^{\dag}$ & $0.951$ & $0.949$ & $0.947$ & $0.945^{\dag}$ & $0.945$ & $\textbf{0.944}$ & $\textbf{0.943}$ & $0.941$\\
& FSDD & $0.922^{\dag}$ & $0.930^{\dag}$ & $0.923^{\dag}$ & $0.919^{\dag}$ & $0.916^{\dag}$ & $0.932^{\dag}$ & $0.934^{\dag}$ & $0.932^{\dag}$ & $0.930^{\dag}$ & $0.931^{\dag}$\\
& SCHI & $\textbf{0.964}^{\ddag}$ & $\textbf{0.957}^{\ddag}$ & $\textbf{0.954}$ & $\textbf{0.950}$ & $\textbf{0.949}$ & $\textbf{0.948}$ & $\textbf{0.947}$ & $\textbf{0.944}$ & $0.942$ & $0.941$\\
& SSU & $0.958^{\dag}$ & $0.953^{\dag}$ & $0.949^{\dag}$ & $0.947^{\dag}$ & $0.946$ & $0.946$ & $0.945$ & $\textbf{0.944}$ & $0.942$ & $0.942$\\
& CHI & $0.948^{\dag}$ & $0.940^{\dag}$ & $0.933^{\dag}$ & $0.929^{\dag}$ & $0.930^{\dag}$ & $0.927^{\dag}$ & $0.942^{\dag}$ & $0.940^{\dag}$ & $0.939^{\dag}$ & $0.939^{\dag}$\\
& SU & $0.951^{\dag}$ & $0.948^{\dag}$ & $0.946^{\dag}$ & $0.946^{\dag}$ & $0.945^{\dag}$ & $0.943^{\dag}$ & $0.942^{\dag}$ & $0.941^{\dag}$ & $0.942^{\dag}$ & $0.942$\\
& ReliefF & $0.918^{\dag}$ & $0.927^{\dag}$ & $0.933^{\dag}$ & $0.937^{\dag}$ & $0.940^{\dag}$ & $0.940^{\dag}$ & $0.941^{\dag}$ & $0.943$ & $\textbf{0.943}$ & $\textbf{0.943}^{\ddag}$\\
\midrule
\multirow{9}{*}{1NN}
& MDFS & 0.924 & \textbf{0.930} & \textbf{0.931} & \textbf{0.932} & \textbf{0.932} & \textbf{0.932} & \textbf{0.932} & \textbf{0.932} & \textbf{0.933} & \textbf{0.933}\\
& MAUCD & $0.894^{\dag}$ & $0.901^{\dag}$ & $0.903^{\dag}$ & $0.902^{\dag}$ & $0.905^{\dag}$ & $0.912^{\dag}$ & $0.912^{\dag}$ & $0.914^{\dag}$ & $0.917^{\dag}$ & $0.920^{\dag}$\\
& mRMR & $\textbf{0.927}^{\ddag}$ & $0.928$ & $0.929^{\dag}$ & $0.929^{\dag}$ & $0.929^{\dag}$ & $0.929^{\dag}$ & $0.929^{\dag}$ & $0.930^{\dag}$ & $0.930^{\dag}$ & $0.930^{\dag}$\\
& FSDD & $0.855^{\dag}$ & $0.883^{\dag}$ & $0.887^{\dag}$ & $0.890^{\dag}$ & $0.895^{\dag}$ & $0.911^{\dag}$ & $0.913^{\dag}$ & $0.913^{\dag}$ & $0.912^{\dag}$ & $0.916^{\dag}$\\
& SCHI & $0.920^{\dag}$ & $0.924^{\dag}$ & $0.924^{\dag}$ & $0.925^{\dag}$ & $0.926^{\dag}$ & $0.927^{\dag}$ & $0.927^{\dag}$ & $0.929^{\dag}$ & $0.931^{\dag}$ & $0.931^{\dag}$\\
& SSU & $0.914^{\dag}$ & $0.919^{\dag}$ & $0.922^{\dag}$ & $0.924^{\dag}$ & $0.927^{\dag}$ & $0.929^{\dag}$ & $0.930^{\dag}$ & $\textbf{0.932}$ & $0.932$ & $\textbf{0.933}$\\
& CHI & $0.891^{\dag}$ & $0.897^{\dag}$ & $0.902^{\dag}$ & $0.903^{\dag}$ & $0.907^{\dag}$ & $0.910^{\dag}$ & $0.922^{\dag}$ & $0.921^{\dag}$ & $0.923^{\dag}$ & $0.924^{\dag}$\\
& SU & $0.905^{\dag}$ & $0.914^{\dag}$ & $0.916^{\dag}$ & $0.919^{\dag}$ & $0.921^{\dag}$ & $0.921^{\dag}$ & $0.923^{\dag}$ & $0.923^{\dag}$ & $0.924^{\dag}$ & $0.925^{\dag}$\\
& ReliefF & $0.865^{\dag}$ & $0.895^{\dag}$ & $0.906^{\dag}$ & $0.913^{\dag}$ & $0.918^{\dag}$ & $0.920^{\dag}$ & $0.922^{\dag}$ & $0.925^{\dag}$ & $0.927^{\dag}$ & $0.927^{\dag}$\\
\midrule
\multirow{9}{*}{SVM}
& MDFS & 0.935 & \textbf{0.941} & \textbf{0.941} & \textbf{0.942} & 0.942 & 0.942 & 0.942 & 0.942 & 0.942 & 0.943\\
& MAUCD & $0.919^{\dag}$ & $0.924^{\dag}$ & $0.928^{\dag}$ & $0.928^{\dag}$ & $0.929^{\dag}$ & $0.935^{\dag}$ & $0.935^{\dag}$ & $0.935^{\dag}$ & $0.939^{\dag}$ & $0.942^{\dag}$\\
& mRMR & $\textbf{0.936}$ & $0.939^{\dag}$ & $0.939^{\dag}$ & $0.940$ & $0.940^{\dag}$ & $0.941^{\dag}$ & $0.941$ & $0.942$ & $0.943$ & $0.943$\\
& FSDD & $0.856^{\dag}$ & $0.905^{\dag}$ & $0.907^{\dag}$ & $0.909^{\dag}$ & $0.911^{\dag}$ & $0.933^{\dag}$ & $0.935^{\dag}$ & $0.935^{\dag}$ & $0.935^{\dag}$ & $0.937^{\dag}$\\
& SCHI & $\textbf{0.936}$ & $0.938^{\dag}$ & $0.938^{\dag}$ & $0.939^{\dag}$ & $0.940^{\dag}$ & $0.940^{\dag}$ & $0.941^{\dag}$ & $0.943^{\ddag}$ & $\textbf{0.945}^{\ddag}$ & $\textbf{0.945}^{\ddag}$\\
& SSU & $0.931^{\dag}$ & $0.937^{\dag}$ & $0.939^{\dag}$ & $\textbf{0.942}$ & $\textbf{0.944}^{\ddag}$ & $\textbf{0.944}^{\ddag}$ & $\textbf{0.944}^{\ddag}$ & $\textbf{0.944}^{\ddag}$ & $0.944^{\ddag}$ & $\textbf{0.945}^{\ddag}$\\
& CHI & $0.910^{\dag}$ & $0.915^{\dag}$ & $0.919^{\dag}$ & $0.922^{\dag}$ & $0.930^{\dag}$ & $0.931^{\dag}$ & $0.937^{\dag}$ & $0.939^{\dag}$ & $0.939^{\dag}$ & $0.941^{\dag}$\\
& SU & $0.927^{\dag}$ & $0.933^{\dag}$ & $0.935^{\dag}$ & $0.937^{\dag}$ & $0.938^{\dag}$ & $0.939^{\dag}$ & $0.940^{\dag}$ & $0.941^{\dag}$ & $0.942$ & $0.943$\\
& ReliefF & $0.905^{\dag}$ & $0.922^{\dag}$ & $0.928^{\dag}$ & $0.932^{\dag}$ & $0.936^{\dag}$ & $0.938^{\dag}$ & $0.939^{\dag}$ & $0.941^{\dag}$ & $0.941^{\dag}$ & $0.942^{\dag}$\\
\bottomrule
\end{tabular}
\end{lrbox}
\scalebox{0.85}{\usebox{\Washington}}
\end{table}
\begin{table}
\centering
\caption{Average MAUC obtained with the nine compared methods on the Indiana data set. The results were obtained by repeating 10-fold cross-validation procedure for 10 times. For each classifier and feature subset size, the Wilcoxon signed-rank test with 95\% confidence level is employed to compare MDFS and 8 other methods. The methods that performed significantly worse (better) than MDFS are highlighted with \dag (\ddag). No superscript is used if no statistical significant difference is detected. The largest MAUC value of each configuration is in boldface.}
\newsavebox{\Indiana}
\begin{lrbox}{\Indiana}
\begin{tabular}{lr*{10}{c}}
\toprule
& & \multicolumn{10}{c}{Feature Subset Size}\\
\cmidrule(r){3-12}
& & 10 & 20 & 30 & 40 & 50 & 60 & 70 & 80 & 90 & 100\\
\midrule
\multirow{9}{*}{Naive Bayes}
& MDFS & 0.890 & 0.892 & 0.893 & 0.891 & 0.890 & 0.890 & 0.890 & 0.890 & 0.890 & 0.890\\
& MAUCD & $0.869^{\dag}$ & $0.874^{\dag}$ & $0.873^{\dag}$ & $0.872^{\dag}$ & $0.872^{\dag}$ & $0.872^{\dag}$ & $0.873^{\dag}$ & $0.874^{\dag}$ & $0.875^{\dag}$ & $0.877^{\dag}$\\
& mRMR & $\textbf{0.894}^{\ddag}$ & $\textbf{0.894}^{\ddag}$ & $\textbf{0.896}^{\ddag}$ & $\textbf{0.896}^{\ddag}$ & $\textbf{0.897}^{\ddag}$ & $\textbf{0.895}^{\ddag}$ & $\textbf{0.894}^{\ddag}$ & $\textbf{0.894}^{\ddag}$ & $\textbf{0.893}^{\ddag}$ & $\textbf{0.893}^{\ddag}$\\
& FSDD & $0.867^{\dag}$ & $0.868^{\dag}$ & $0.865^{\dag}$ & $0.865^{\dag}$ & $0.866^{\dag}$ & $0.869^{\dag}$ & $0.871^{\dag}$ & $0.873^{\dag}$ & $0.875^{\dag}$ & $0.877^{\dag}$\\
& SCHI & $0.888^{\dag}$ & $0.888^{\dag}$ & $0.888^{\dag}$ & $0.886^{\dag}$ & $0.886^{\dag}$ & $0.885^{\dag}$ & $0.885^{\dag}$ & $0.885^{\dag}$ & $0.884^{\dag}$ & $0.884^{\dag}$\\
& SSU & $0.887^{\dag}$ & $0.886^{\dag}$ & $0.885^{\dag}$ & $0.884^{\dag}$ & $0.884^{\dag}$ & $0.883^{\dag}$ & $0.883^{\dag}$ & $0.883^{\dag}$ & $0.883^{\dag}$ & $0.883^{\dag}$\\
& CHI & $0.867^{\dag}$ & $0.866^{\dag}$ & $0.864^{\dag}$ & $0.868^{\dag}$ & $0.872^{\dag}$ & $0.874^{\dag}$ & $0.876^{\dag}$ & $0.877^{\dag}$ & $0.878^{\dag}$ & $0.878^{\dag}$\\
& SU & $0.868^{\dag}$ & $0.867^{\dag}$ & $0.865^{\dag}$ & $0.866^{\dag}$ & $0.870^{\dag}$ & $0.872^{\dag}$ & $0.874^{\dag}$ & $0.876^{\dag}$ & $0.877^{\dag}$ & $0.878^{\dag}$\\
& ReliefF & $0.874^{\dag}$ & $0.882^{\dag}$ & $0.883^{\dag}$ & $0.882^{\dag}$ & $0.882^{\dag}$ & $0.882^{\dag}$ & $0.881^{\dag}$ & $0.881^{\dag}$ & $0.881^{\dag}$ & $0.882^{\dag}$\\
\midrule
\multirow{9}{*}{C4.5}
& MDFS & 0.882 & 0.889 & 0.892 & 0.893 & 0.893 & 0.893 & 0.891 & 0.893 & 0.892 & 0.891\\
& MAUCD & $0.797^{\dag}$ & $0.847^{\dag}$ & $0.846^{\dag}$ & $0.847^{\dag}$ & $0.844^{\dag}$ & $0.842^{\dag}$ & $0.852^{\dag}$ & $0.851^{\dag}$ & $0.868^{\dag}$ & $0.876^{\dag}$\\
& mRMR & $0.866^{\dag}$ & $0.870^{\dag}$ & $0.876^{\dag}$ & $0.882^{\dag}$ & $0.882^{\dag}$ & $0.886^{\dag}$ & $0.888^{\dag}$ & $0.888^{\dag}$ & $0.888^{\dag}$ & $0.889$\\
& FSDD & $0.804^{\dag}$ & $0.805^{\dag}$ & $0.808^{\dag}$ & $0.817^{\dag}$ & $0.840^{\dag}$ & $0.850^{\dag}$ & $0.851^{\dag}$ & $0.849^{\dag}$ & $0.847^{\dag}$ & $0.868^{\dag}$\\
& SCHI & $\textbf{0.904}^{\ddag}$ & $\textbf{0.904}^{\ddag}$ & $\textbf{0.902}^{\ddag}$ & $\textbf{0.902}^{\ddag}$ & $0.900^{\ddag}$ & $0.899^{\ddag}$ & $\textbf{0.900}^{\ddag}$ & $\textbf{0.899}^{\ddag}$ & $\textbf{0.900}^{\ddag}$ & $\textbf{0.899}^{\ddag}$\\
& SSU & $0.895^{\ddag}$ & $0.901^{\ddag}$ & $\textbf{0.902}^{\ddag}$ & $\textbf{0.902}^{\ddag}$ & $\textbf{0.902}^{\ddag}$ & $\textbf{0.900}^{\ddag}$ & $\textbf{0.900}^{\ddag}$ & $\textbf{0.899}^{\ddag}$ & $0.899^{\ddag}$ & $0.898^{\ddag}$\\
& CHI & $0.797^{\dag}$ & $0.811^{\dag}$ & $0.823^{\dag}$ & $0.854^{\dag}$ & $0.870^{\dag}$ & $0.877^{\dag}$ & $0.879^{\dag}$ & $0.881^{\dag}$ & $0.881^{\dag}$ & $0.881^{\dag}$\\
& SU & $0.810^{\dag}$ & $0.808^{\dag}$ & $0.819^{\dag}$ & $0.845^{\dag}$ & $0.855^{\dag}$ & $0.859^{\dag}$ & $0.874^{\dag}$ & $0.875^{\dag}$ & $0.877^{\dag}$ & $0.878^{\dag}$\\
& ReliefF & $0.860^{\dag}$ & $0.874^{\dag}$ & $0.884^{\dag}$ & $0.887^{\dag}$ & $0.889^{\dag}$ & $0.889^{\dag}$ & $0.889^{\dag}$ & $0.890$ & $0.890$ & $0.891$\\
\midrule
\multirow{9}{*}{1NN}
& MDFS & 0.864 & 0.896 & 0.909 & 0.917 & 0.919 & 0.921 & 0.923 & 0.924 & 0.923 & 0.922\\
& MAUCD & $0.759^{\dag}$ & $0.852^{\dag}$ & $0.876^{\dag}$ & $0.875^{\dag}$ & $0.874^{\dag}$ & $0.855^{\dag}$ & $0.868^{\dag}$ & $0.872^{\dag}$ & $0.883^{\dag}$ & $0.904^{\dag}$\\
& mRMR & $0.815^{\dag}$ & $0.826^{\dag}$ & $0.841^{\dag}$ & $0.863^{\dag}$ & $0.875^{\dag}$ & $0.882^{\dag}$ & $0.887^{\dag}$ & $0.894^{\dag}$ & $0.896^{\dag}$ & $0.898^{\dag}$\\
& FSDD & $0.752^{\dag}$ & $0.796^{\dag}$ & $0.819^{\dag}$ & $0.841^{\dag}$ & $0.860^{\dag}$ & $0.870^{\dag}$ & $0.872^{\dag}$ & $0.869^{\dag}$ & $0.854^{\dag}$ & $0.886^{\dag}$\\
& SCHI & $0.877^{\ddag}$ & $0.891^{\dag}$ & $0.905^{\dag}$ & $0.911^{\dag}$ & $0.915$ & $0.915^{\dag}$ & $0.916^{\dag}$ & $0.918^{\dag}$ & $0.916^{\dag}$ & $0.915^{\dag}$\\
& SSU & $\textbf{0.897}^{\ddag}$ & $\textbf{0.922}^{\ddag}$ & $\textbf{0.931}^{\ddag}$ & $\textbf{0.935}^{\ddag}$ & $\textbf{0.938}^{\ddag}$ & $\textbf{0.939}^{\ddag}$ & $\textbf{0.940}^{\ddag}$ & $\textbf{0.940}^{\ddag}$ & $\textbf{0.940}^{\ddag}$ & $\textbf{0.940}^{\ddag}$\\
& CHI & $0.762^{\dag}$ & $0.820^{\dag}$ & $0.847^{\dag}$ & $0.898^{\dag}$ & $0.915^{\dag}$ & $0.924^{\ddag}$ & $0.927^{\ddag}$ & $0.927^{\ddag}$ & $0.929^{\ddag}$ & $0.928^{\ddag}$\\
& SU & $0.767^{\dag}$ & $0.798^{\dag}$ & $0.827^{\dag}$ & $0.863^{\dag}$ & $0.877^{\dag}$ & $0.886^{\dag}$ & $0.907^{\dag}$ & $0.912^{\dag}$ & $0.914^{\dag}$ & $0.917^{\dag}$\\
& ReliefF & $0.866$ & $0.897$ & $0.913^{\ddag}$ & $0.919$ & $0.922^{\ddag}$ & $0.925^{\ddag}$ & $0.925^{\ddag}$ & $0.926^{\ddag}$ & $0.927^{\ddag}$ & $0.927^{\ddag}$\\
\midrule
\multirow{9}{*}{SVM}
& MDFS & 0.890 & 0.927 & 0.941 & 0.947 & 0.950 & 0.952 & 0.954 & 0.955 & 0.957 & 0.957\\
& MAUCD & $0.774^{\dag}$ & $0.867^{\dag}$ & $0.882^{\dag}$ & $0.887^{\dag}$ & $0.891^{\dag}$ & $0.894^{\dag}$ & $0.918^{\dag}$ & $0.921^{\dag}$ & $0.948^{\dag}$ & $0.953^{\dag}$\\
& mRMR & $0.888$ & $0.915^{\dag}$ & $0.931^{\dag}$ & $0.938^{\dag}$ & $0.942^{\dag}$ & $0.946^{\dag}$ & $0.950^{\dag}$ & $0.953^{\dag}$ & $0.955^{\dag}$ & $0.957$\\
& FSDD & $0.786^{\dag}$ & $0.822^{\dag}$ & $0.836^{\dag}$ & $0.863^{\dag}$ & $0.894^{\dag}$ & $0.910^{\dag}$ & $0.916^{\dag}$ & $0.920^{\dag}$ & $0.921^{\dag}$ & $0.952^{\dag}$\\
& SCHI & $0.915^{\ddag}$ & $0.936^{\ddag}$ & $0.943^{\ddag}$ & $0.946$ & $0.948^{\dag}$ & $0.949^{\dag}$ & $0.953$ & $0.955$ & $0.956$ & $0.958$\\
& SSU & $\textbf{0.927}^{\ddag}$ & $\textbf{0.949}^{\ddag}$ & $\textbf{0.954}^{\ddag}$ & $\textbf{0.957}^{\ddag}$ & $\textbf{0.959}^{\ddag}$ & $\textbf{0.961}^{\ddag}$ & $\textbf{0.961}^{\ddag}$ & $\textbf{0.962}^{\ddag}$ & $\textbf{0.963}^{\ddag}$ & $\textbf{0.964}^{\ddag}$\\
& CHI & $0.778^{\dag}$ & $0.833^{\dag}$ & $0.864^{\dag}$ & $0.906^{\dag}$ & $0.937^{\dag}$ & $0.945^{\dag}$ & $0.950^{\dag}$ & $0.953^{\dag}$ & $0.952^{\dag}$ & $0.953^{\dag}$\\
& SU & $0.784^{\dag}$ & $0.818^{\dag}$ & $0.848^{\dag}$ & $0.887^{\dag}$ & $0.912^{\dag}$ & $0.926^{\dag}$ & $0.947^{\dag}$ & $0.950^{\dag}$ & $0.952^{\dag}$ & $0.954^{\dag}$\\
& ReliefF & $0.882^{\dag}$ & $0.911^{\dag}$ & $0.930^{\dag}$ & $0.940^{\dag}$ & $0.946^{\dag}$ & $0.950^{\dag}$ & $0.953$ & $0.955$ & $0.956$ & $0.957$\\
\bottomrule
\end{tabular}
\end{lrbox}
\scalebox{0.85}{\usebox{\Indiana}}
\end{table}
\begin{table}
\centering
\caption{Average MAUC obtained with the nine compared methods on the Synthetic data set. The results were obtained by repeating 10-fold cross-validation procedure for 10 times. For each classifier and feature subset size, the Wilcoxon signed-rank test with 95\% confidence level is employed to compare MDFS and 8 other methods. The methods that performed significantly worse (better) than MDFS are highlighted with \dag (\ddag). No superscript is used if no statistical significant difference is detected. The largest MAUC value of each configuration is in boldface.}
\begin{tabular}{lr*{6}{c}}
\toprule
& & \multicolumn{6}{c}{Feature Subset Size}\\
\cmidrule(r){3-8}
& & 10 & 20 & 30 & 40 & 50 & 60\\
\midrule
\multirow{9}{*}{Naive Bayes}
& MDFS & \textbf{0.992} & \textbf{0.996} & \textbf{0.997} & \textbf{0.998} & 0.998 & \textbf{0.998}\\
& MAUCD & $0.971^{\dag}$ & $0.968^{\dag}$ & $0.980^{\dag}$ & $0.993^{\dag}$ & $0.998^{\dag}$ & $\textbf{0.998}$\\
& mRMR & $\textbf{0.992}$ & $0.993^{\dag}$ & $0.995^{\dag}$ & $0.997^{\dag}$ & $0.998^{\dag}$ & $\textbf{0.998}$\\
& FSDD & $0.971^{\dag}$ & $0.968^{\dag}$ & $0.979^{\dag}$ & $0.991^{\dag}$ & $0.997^{\dag}$ & $\textbf{0.998}$\\
& SCHI & $0.984^{\dag}$ & $0.988^{\dag}$ & $0.990^{\dag}$ & $0.997^{\dag}$ & $\textbf{0.999}^{\ddag}$ & $\textbf{0.998}$\\
& SSU & $0.975^{\dag}$ & $0.979^{\dag}$ & $0.985^{\dag}$ & $0.994^{\dag}$ & $0.998^{\dag}$ & $\textbf{0.998}$\\
& CHI & $0.971^{\dag}$ & $0.968^{\dag}$ & $0.982^{\dag}$ & $0.996^{\dag}$ & $0.998^{\dag}$ & $\textbf{0.998}$\\
& SU & $0.969^{\dag}$ & $0.968^{\dag}$ & $0.978^{\dag}$ & $0.992^{\dag}$ & $0.998^{\dag}$ & $\textbf{0.998}$\\
& ReliefF & $0.968^{\dag}$ & $0.970^{\dag}$ & $0.981^{\dag}$ & $0.993^{\dag}$ & $0.997^{\dag}$ & $\textbf{0.998}$\\
\midrule
\multirow{9}{*}{C4.5}
& MDFS & 0.950 & 0.959 & 0.963 & 0.970 & 0.968 & \textbf{0.966}\\
& MAUCD & $0.929^{\dag}$ & $0.909^{\dag}$ & $0.927^{\dag}$ & $0.949^{\dag}$ & $0.967$ & $\textbf{0.966}$\\
& mRMR & $\textbf{0.964}^{\ddag}$ & $\textbf{0.966}^{\ddag}$ & $\textbf{0.972}^{\ddag}$ & $\textbf{0.972}$ & $0.969$ & $\textbf{0.966}$\\
& FSDD & $0.928^{\dag}$ & $0.905^{\dag}$ & $0.926^{\dag}$ & $0.952^{\dag}$ & $0.961^{\dag}$ & $\textbf{0.966}$\\
& SCHI & $0.943$ & $0.949^{\dag}$ & $0.954^{\dag}$ & $0.966$ & $0.966^{\dag}$ & $\textbf{0.966}$\\
& SSU & $0.925^{\dag}$ & $0.932^{\dag}$ & $0.954^{\dag}$ & $0.966$ & $0.967$ & $\textbf{0.966}$\\
& CHI & $0.933^{\dag}$ & $0.910^{\dag}$ & $0.937^{\dag}$ & $0.966$ & $\textbf{0.972}^{\ddag}$ & $\textbf{0.966}$\\
& SU & $0.928^{\dag}$ & $0.907^{\dag}$ & $0.925^{\dag}$ & $0.954^{\dag}$ & $0.965^{\dag}$ & $\textbf{0.966}$\\
& ReliefF & $0.921^{\dag}$ & $0.916^{\dag}$ & $0.932^{\dag}$ & $0.954^{\dag}$ & $0.965^{\dag}$ & $\textbf{0.966}$\\
\midrule
\multirow{9}{*}{1NN}
& MDFS & 0.929 & \textbf{0.965} & \textbf{0.979} & \textbf{0.987} & 0.989 & \textbf{0.982}\\
& MAUCD & $0.880^{\dag}$ & $0.919^{\dag}$ & $0.956^{\dag}$ & $0.983^{\dag}$ & $\textbf{0.991}$ & $\textbf{0.982}$\\
& mRMR & $\textbf{0.942}^{\ddag}$ & $0.956^{\dag}$ & $0.977$ & $0.984^{\dag}$ & $0.987^{\dag}$ & $\textbf{0.982}$\\
& FSDD & $0.887^{\dag}$ & $0.910^{\dag}$ & $0.941^{\dag}$ & $0.967^{\dag}$ & $0.984^{\dag}$ & $\textbf{0.982}$\\
& SCHI & $0.896^{\dag}$ & $0.952^{\dag}$ & $0.970^{\dag}$ & $0.981^{\dag}$ & $0.986^{\dag}$ & $\textbf{0.982}$\\
& SSU & $0.887^{\dag}$ & $0.923^{\dag}$ & $0.950^{\dag}$ & $0.973^{\dag}$ & $0.986^{\dag}$ & $\textbf{0.982}$\\
& CHI & $0.876^{\dag}$ & $0.917^{\dag}$ & $0.953^{\dag}$ & $0.985$ & $0.988$ & $\textbf{0.982}$\\
& SU & $0.891^{\dag}$ & $0.914^{\dag}$ & $0.946^{\dag}$ & $0.978^{\dag}$ & $0.985^{\dag}$ & $\textbf{0.982}$\\
& ReliefF & $0.885^{\dag}$ & $0.920^{\dag}$ & $0.955^{\dag}$ & $0.977^{\dag}$ & $0.985^{\dag}$ & $\textbf{0.982}$\\
\midrule
\multirow{9}{*}{SVM}
& MDFS & 0.946 & \textbf{0.984} & \textbf{0.994} & 0.996 & 0.997 & \textbf{0.999}\\
& MAUCD & $0.880^{\dag}$ & $0.945^{\dag}$ & $0.979^{\dag}$ & $0.995$ & $0.998^{\ddag}$ & $\textbf{0.999}$\\
& mRMR & $\textbf{0.947}$ & $0.980^{\dag}$ & $\textbf{0.994}$ & $0.996$ & $0.997$ & $\textbf{0.999}$\\
& FSDD & $0.881^{\dag}$ & $0.942^{\dag}$ & $0.972^{\dag}$ & $0.993^{\dag}$ & $0.998$ & $\textbf{0.999}$\\
& SCHI & $0.919^{\dag}$ & $0.969^{\dag}$ & $0.990^{\dag}$ & $0.995^{\dag}$ & $0.997$ & $\textbf{0.999}$\\
& SSU & $0.898^{\dag}$ & $0.959^{\dag}$ & $0.973^{\dag}$ & $0.994^{\dag}$ & $0.998^{\ddag}$ & $\textbf{0.999}$\\
& CHI & $0.894^{\dag}$ & $0.952^{\dag}$ & $0.977^{\dag}$ & $\textbf{0.997}$ & $0.998^{\ddag}$ & $\textbf{0.999}$\\
& SU & $0.934^{\dag}$ & $0.956^{\dag}$ & $0.979^{\dag}$ & $0.993^{\dag}$ & $\textbf{0.999}^{\ddag}$ & $\textbf{0.999}$\\
& ReliefF & $0.925^{\dag}$ & $0.959^{\dag}$ & $0.982^{\dag}$ & $0.994^{\dag}$ & $0.997$ & $\textbf{0.999}$\\
\bottomrule
\end{tabular}
\end{table}
\begin{table}
\caption{Average MAUC obtained with the nine compared methods on the Thyroid data set. The results were obtained by repeating 10-fold cross-validation procedure for 10 times. For each classifier and feature subset size, the Wilcoxon signed-rank test with 95\% confidence level is employed to compare MDFS and 8 other methods. The methods that performed significantly worse (better) than MDFS are highlighted with \dag (\ddag). No superscript is used if no statistical significant difference is detected. The largest MAUC value of each configuration is in boldface.}
\newsavebox{\Thyroid}
\begin{lrbox}{\Thyroid}
\begin{tabular}{lr*{10}{c}}
\toprule
& & \multicolumn{10}{c}{Feature Subset Size}\\
\cmidrule(r){3-12}
& & 10 & 20 & 30 & 40 & 50 & 60 & 70 & 80 & 90 & 100\\
\midrule
\multirow{9}{*}{Naive Bayes}
& MDFS & 0.857 & 0.889 & 0.905 & \textbf{0.911} & \textbf{0.911} & 0.910 & 0.906 & 0.904 & 0.902 & 0.901\\
& MAUCD & $0.879^{\ddag}$ & $0.890$ & $0.895^{\dag}$ & $0.900^{\dag}$ & $0.906^{\dag}$ & $0.906^{\dag}$ & $0.906$ & $0.905$ & $0.903$ & $0.900$\\
& mRMR & $\textbf{0.882}^{\ddag}$ & $\textbf{0.893}$ & $0.899^{\dag}$ & $0.903^{\dag}$ & $0.908$ & $0.907$ & $0.906$ & $0.906$ & $0.902$ & $0.903$\\
& FSDD & $0.871^{\ddag}$ & $0.892$ & $\textbf{0.909}$ & $0.910$ & $0.909$ & $\textbf{0.911}$ & $\textbf{0.911}$ & $\textbf{0.913}^{\ddag}$ & $\textbf{0.914}^{\ddag}$ & $\textbf{0.913}^{\ddag}$\\
& SCHI & $0.872^{\ddag}$ & $0.881^{\dag}$ & $0.885^{\dag}$ & $0.888^{\dag}$ & $0.891^{\dag}$ & $0.891^{\dag}$ & $0.894^{\dag}$ & $0.898^{\dag}$ & $0.900$ & $0.902$\\
& SSU & $0.874^{\ddag}$ & $0.880$ & $0.889^{\dag}$ & $0.890^{\dag}$ & $0.893^{\dag}$ & $0.895^{\dag}$ & $0.898^{\dag}$ & $0.899$ & $0.903$ & $0.904$\\
& CHI & $0.860$ & $0.882$ & $0.893^{\dag}$ & $0.902^{\dag}$ & $0.905^{\dag}$ & $0.909$ & $0.909$ & $0.909$ & $0.907$ & $0.904$\\
& SU & $0.868$ & $0.882$ & $0.892^{\dag}$ & $0.898^{\dag}$ & $0.905$ & $0.908$ & $0.909$ & $0.908$ & $0.906$ & $0.903$\\
& ReliefF & $0.855$ & $0.882$ & $0.891^{\dag}$ & $0.895^{\dag}$ & $0.900^{\dag}$ & $0.901^{\dag}$ & $0.901^{\dag}$ & $0.900$ & $0.901$ & $0.903$\\
\midrule
\multirow{9}{*}{C4.5}
& MDFS & 0.746 & 0.759 & 0.757 & 0.778 & \textbf{0.775} & \textbf{0.769} & 0.760 & \textbf{0.762} & 0.753 & 0.748\\
& MAUCD & $0.747$ & $0.763$ & $0.776$ & $0.777$ & $0.768$ & $0.760$ & $0.756$ & $0.754$ & $0.751$ & $0.748$\\
& mRMR & $0.767$ & $0.769$ & $0.750$ & $0.742^{\dag}$ & $0.748^{\dag}$ & $0.750$ & $0.740^{\dag}$ & $0.739^{\dag}$ & $0.744$ & $0.743$\\
& FSDD & $0.775^{\ddag}$ & $0.763$ & $0.769$ & $0.766$ & $0.744^{\dag}$ & $0.738^{\dag}$ & $0.759$ & $0.753$ & $0.746$ & $0.744$\\
& SCHI & $0.807^{\ddag}$ & $0.770$ & $0.767$ & $0.768$ & $0.772$ & $0.761$ & $\textbf{0.765}$ & $0.749$ & $0.751$ & $0.749$\\
& SSU & $\textbf{0.816}^{\ddag}$ & $0.781^{\ddag}$ & $0.772$ & $0.767$ & $0.760$ & $0.762$ & $0.759$ & $\textbf{0.762}$ & $\textbf{0.760}$ & $\textbf{0.752}$\\
& CHI & $0.795^{\ddag}$ & $\textbf{0.784}^{\ddag}$ & $\textbf{0.788}^{\ddag}$ & $0.783$ & $0.768$ & $0.758$ & $0.756$ & $0.751$ & $0.751$ & $0.748$\\
& SU & $0.764$ & $0.767$ & $0.781^{\ddag}$ & $\textbf{0.785}$ & $0.772$ & $0.764$ & $0.758$ & $0.751$ & $0.746$ & $0.749$\\
& ReliefF & $0.719^{\dag}$ & $0.738$ & $0.738^{\dag}$ & $0.740^{\dag}$ & $0.736^{\dag}$ & $0.736^{\dag}$ & $0.742$ & $0.739^{\dag}$ & $0.738$ & $0.748$\\
\midrule
\multirow{9}{*}{1NN}
& MDFS & 0.727 & 0.762 & \textbf{0.784} & \textbf{0.806} & \textbf{0.819} & \textbf{0.827} & \textbf{0.817} & \textbf{0.820} & 0.815 & 0.818\\
& MAUCD & $0.753^{\ddag}$ & $\textbf{0.763}$ & $0.770$ & $0.785^{\dag}$ & $0.798^{\dag}$ & $0.799^{\dag}$ & $0.814$ & $0.808^{\dag}$ & $0.812$ & $0.815$\\
& mRMR & $0.753^{\ddag}$ & $0.756$ & $0.767$ & $0.792^{\dag}$ & $0.797^{\dag}$ & $0.805^{\dag}$ & $0.815$ & $0.819$ & $0.815$ & $0.814$\\
& FSDD & $0.748$ & $0.741^{\dag}$ & $0.771$ & $0.773^{\dag}$ & $0.797^{\dag}$ & $0.810^{\dag}$ & $0.814$ & $0.813$ & $\textbf{0.821}$ & $\textbf{0.820}$\\
& SCHI & $0.738$ & $0.755$ & $0.761^{\dag}$ & $0.760^{\dag}$ & $0.760^{\dag}$ & $0.752^{\dag}$ & $0.769^{\dag}$ & $0.768^{\dag}$ & $0.780^{\dag}$ & $0.787^{\dag}$\\
& SSU & $0.747^{\ddag}$ & $0.760$ & $0.758^{\dag}$ & $0.764^{\dag}$ & $0.775^{\dag}$ & $0.778^{\dag}$ & $0.778^{\dag}$ & $0.777^{\dag}$ & $0.798^{\dag}$ & $0.797^{\dag}$\\
& CHI & $0.748^{\ddag}$ & $0.743$ & $0.769$ & $0.772^{\dag}$ & $0.781^{\dag}$ & $0.790^{\dag}$ & $0.799^{\dag}$ & $0.798^{\dag}$ & $0.799^{\dag}$ & $0.802$\\
& SU & $\textbf{0.762}^{\ddag}$ & $0.749$ & $0.761^{\dag}$ & $0.772^{\dag}$ & $0.785^{\dag}$ & $0.799^{\dag}$ & $0.798^{\dag}$ & $0.800^{\dag}$ & $0.801$ & $0.796^{\dag}$\\
& ReliefF & $0.718$ & $0.750$ & $0.765$ & $0.774^{\dag}$ & $0.769^{\dag}$ & $0.773^{\dag}$ & $0.785^{\dag}$ & $0.787^{\dag}$ & $0.788^{\dag}$ & $0.786^{\dag}$\\
\midrule
\multirow{9}{*}{SVM}
& MDFS & 0.758 & 0.792 & \textbf{0.815} & \textbf{0.827} & \textbf{0.828} & \textbf{0.832} & 0.827 & \textbf{0.823} & 0.819 & 0.825\\
& MAUCD & $0.785^{\ddag}$ & $0.794$ & $0.775^{\dag}$ & $0.784^{\dag}$ & $0.798^{\dag}$ & $0.800^{\dag}$ & $0.800^{\dag}$ & $0.810$ & $0.814$ & $0.820$\\
& mRMR & $0.801^{\ddag}$ & $0.790$ & $0.787^{\dag}$ & $0.793^{\dag}$ & $0.811^{\dag}$ & $0.823$ & $\textbf{0.830}$ & $\textbf{0.823}$ & $\textbf{0.830}$ & $\textbf{0.837}$\\
& FSDD & $0.797^{\ddag}$ & $0.783$ & $0.784^{\dag}$ & $0.788^{\dag}$ & $0.792^{\dag}$ & $0.796^{\dag}$ & $0.806^{\dag}$ & $0.821$ & $\textbf{0.830}$ & $0.834$\\
& SCHI & $\textbf{0.804}^{\ddag}$ & $0.801$ & $0.783^{\dag}$ & $0.791^{\dag}$ & $0.795^{\dag}$ & $0.806^{\dag}$ & $0.795^{\dag}$ & $0.800^{\dag}$ & $0.804$ & $0.817$\\
& SSU & $0.797^{\ddag}$ & $0.798$ & $0.784^{\dag}$ & $0.795^{\dag}$ & $0.790^{\dag}$ & $0.799^{\dag}$ & $0.806^{\dag}$ & $0.816$ & $0.820$ & $0.834$\\
& CHI & $0.795^{\ddag}$ & $0.808^{\ddag}$ & $0.796^{\dag}$ & $0.795^{\dag}$ & $0.800^{\dag}$ & $0.804^{\dag}$ & $0.810^{\dag}$ & $0.819$ & $0.809$ & $0.811^{\dag}$\\
& SU & $0.801^{\ddag}$ & $\textbf{0.818}^{\ddag}$ & $0.802$ & $0.805^{\dag}$ & $0.804^{\dag}$ & $0.806^{\dag}$ & $0.821$ & $0.815$ & $0.822$ & $0.827$\\
& ReliefF & $0.762$ & $0.781$ & $0.781^{\dag}$ & $0.782^{\dag}$ & $0.789^{\dag}$ & $0.795^{\dag}$ & $0.794^{\dag}$ & $0.800^{\dag}$ & $0.806$ & $0.805^{\dag}$\\
\bottomrule
\end{tabular}
\end{lrbox}
\scalebox{0.85}{\usebox{\Thyroid}}
\end{table}
    \begin{figure*}
      \centering
      \includegraphics[width=1.0\textwidth]{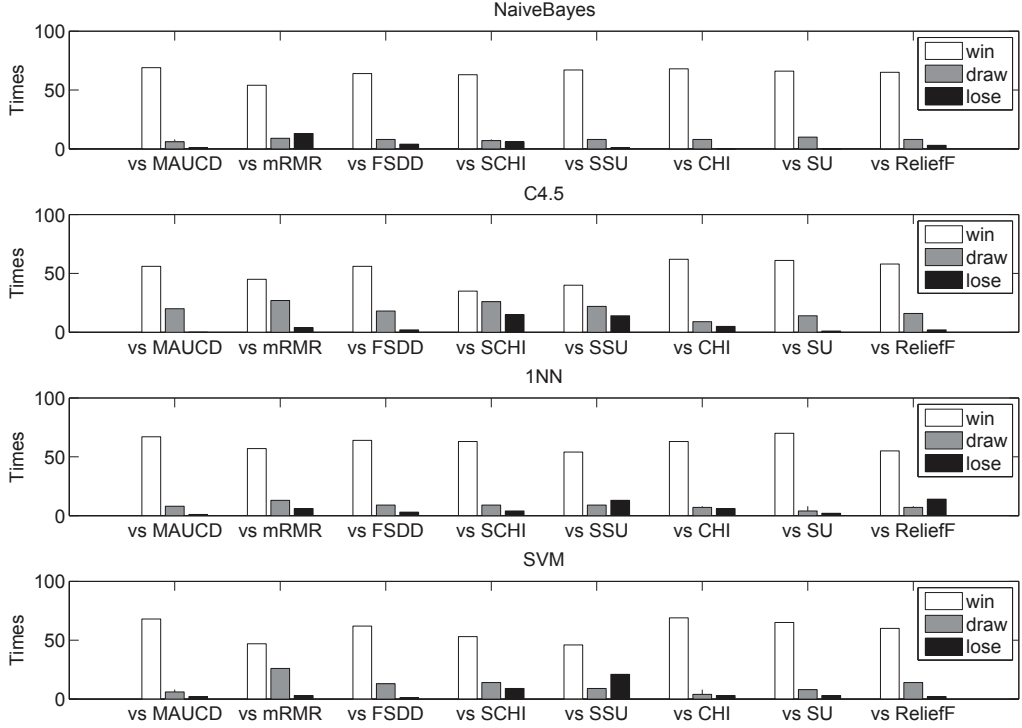}
      \caption{Results of the Wilcoxon signed rank test with 95\% confidence level between MDFS and the 8 other compared feature selection methods on 4 different types of classifiers. The hight of each bar indicates the times that MDFS win (draw or lose) against the counterpart feature selection method over all feature subset sizes and data sets.}
    \end{figure*}
    \begin{table}
      \caption{Average runtime (in seconds) of the compared methods on the 8 data sets.}
      \centering
      \newsavebox{\fstime}
      \begin{lrbox}{\fstime}
      \begin{tabular}{r*{8}{c}}
        \toprule
        & \multicolumn{8}{c}{Data Set}\\
        \cmidrule(r){2-9}
        & ISOLET & MNIST & USPS & Phoneme & Washington & Indiana & Synthetic & Thyroid\\
        \midrule
        MDFS    & 38.815 & 18.765 & 6.223 & 1.778 & 5.814 & 6.351 & 0.054 & 0.268\\
        MAUCD   & 43.109 & 20.949 & 6.324 & 2.161 & 6.324 & 7.090 & 0.063 & 0.323\\
        mRMR    & 773.153 & 716.781 & 114.117 & 60.939 & 204.143 & 153.125 & 0.578 & 22.945\\
        FSDD    & 0.512 & 1.363 & 0.241 & 0.100 & 0.293 & 0.187 & 0.002 & 0.036\\
        SCHI    & 170.276 & 76.982 & 33.007 & 10.615 & 26.889 & 24.107 & 0.282 & 1.000\\
        SSU     & 174.870 & 82.301 & 34.468 & 10.603 & 27.745 & 24.539 & 0.273 & 0.973\\
        CHI     & 18.624 & 8.068 & 2.796 & 3.417 & 5.112 & 3.839 & 0.100 & 0.293\\
        SU      & 19.072 & 8.242 & 2.964 & 3.374 & 5.189 & 3.945 & 0.091 & 0.285\\
        ReliefF & 3.434 & 5.706 & 1.774 & 0.740 & 1.748 & 1.551 & 0.037 & 0.285\\
        \bottomrule
      \end{tabular}
      \end{lrbox}
      \scalebox{0.75}{\usebox{\fstime}}
    \end{table}
  \section{Conclusions and discussions}
    Although numerous successful feature selection methods have been developed for accuracy-oriented classification systems, recent studies revealed that accuracy itself is not an appropriate performance metric in many real-world practices and may lead to undesirable classification systems. Instead, AUC and MAUC are adopted more and more commonly in the literature. This shift of performance metric raises the need for new feature selection methods. In this study, we proposed the MDFS feature selection method for MAUC-oriented classification systems. It was inspired by the observation that MAUC value of a classification system is actually the average of its AUC values on every binary sub-problem that consists of a pair of classes. Therefore, MDFS first decomposes a multi-class problem to a batch of binary class sub-problems in one-versus-one manner. Then, features are iteratively selected based on their utility on each sub-problem. Equal focus on every sub-problem is implemented by choosing one of them with equal probability in each iteration.

    The advantage of MDFS over traditional filter methods has been justified by comparative studies on 8 benchmark data sets. Results obtained with 4 types of classifiers demonstrated that MDFS is overall superior to the 8 other compared filter methods in terms of the MAUC values of classification systems. Experimental studies also showed that the direct use of MAUC as feature ranking metric led to inferior performance compared to MDFS. Finally, the computational complexity of MDFS is comparable to that of most compared filter feature selection methods.

    MDFS might be further improved from two aspects: First, the employment of a random strategy by MDFS in selecting sub-problems does not take the possible correlation between sub-problems into account. If a number of sub-problems are highly correlated with one another, many more features would be selected for them than for the other sub-problems. This will lead to redundant features and make the weight of each problem not equal in feature selection, which may lead to inferior performance of MAUC-oriented classification systems. Hence, finding a way to detect this correlation among sub-problems or the relative importance of different sub-problems in maximizing MAUC is of great interest. Second, redundancy among features has not been considered in MDFS. Having the great success of some redundancy-exclusive strategies \cite{Peng2005} in mind, incorporating them into  MDFS may promise enhanced performance. We will investigate these two issues in the future.

    \section*{Acknowledgements}
    This work was supported in part by the National Natural Science Foundation of China under Grant 60802036 and Grant U0835002. The authors thank Dr. Thomas Weise for proofreading the manuscript.

    \bibliographystyle{model1-num-names}
    \bibliography{MyCollection}
\end{document}